\documentclass[lettersize,journal]{IEEEtran}
\usepackage{amsmath,amsfonts}
\usepackage{algorithmic}
\usepackage{algorithm}
\usepackage{array}
\usepackage[caption=false,font=normalsize,labelfont=sf,textfont=sf]{subfig}
\usepackage{textcomp}
\usepackage{stfloats}
\usepackage{url}
\usepackage{verbatim}
\usepackage{graphicx}
\usepackage{cite}

\usepackage{makecell}
\usepackage{booktabs}
\usepackage{threeparttable}
\usepackage{multirow}
\usepackage{xcolor}
\usepackage{listings}
\begin{document}

\title{Robust Identity Perceptual Watermark Against Deepfake Face Swapping}

\author{Tianyi Wang,~\IEEEmembership{Member,~IEEE,}
        Mengxiao Huang,
        Harry Cheng,~\IEEEmembership{Member,~IEEE,}
        Bin Ma,~\IEEEmembership{Member,~IEEE,}
        and Yinglong Wang,~\IEEEmembership{Senior Member,~IEEE}
\thanks{Tianyi Wang and Harry Cheng are with the School of Computing, National University of Singapore, Singapore (e-mail: wangty@nus.edu.sg and xaCheng1996@gmail.com).}
\thanks{Mengxiao Huang is with Shandong Artificial Intelligence Institute, Qilu University of Technology (Shandong Academy of Sciences), Jinan, China (e-mail: huangmengxiao1@gmail.com).}
\thanks{Bin Ma is with the School of Cyber Security, Qilu University of Technology (Shandong Academy of Sciences), Jinan, China (e-mail: sddxmb@126.com).}
\thanks{Yinglong Wang is with the Key Laboratory of Computing Power Network and Information Security, Ministry of Education, China (e-mail: wangyinglong@qlu.edu.cn).}
\thanks{Corresponding author: Yinglong Wang.}
}



\maketitle

\begin{abstract}
Notwithstanding offering convenience and entertainment to society, Deepfake face swapping has caused critical privacy issues with the rapid development of deep generative models. Due to imperceptible artifacts in high-quality synthetic images, passive detection models against face swapping in recent years usually suffer performance damping regarding the generalizability issue in cross-domain scenarios. Therefore, several studies have been attempted to proactively protect the original images against malicious manipulations by inserting invisible signals in advance. However, existing proactive defense approaches demonstrate unsatisfactory results with respect to visual quality, detection accuracy, and source tracing ability. In this study, to fulfill the research gap, we propose a robust identity perceptual watermarking framework that concurrently performs detection and source tracing against Deepfake face swapping proactively. We innovatively assign identity semantics regarding the image contents to the watermarks and devise an unpredictable and nonreversible chaotic encryption system to ensure watermark confidentiality. The watermarks are robustly encoded and recovered by jointly training an encoder-decoder framework along with adversarial image manipulations. For a suspect image, falsification is accomplished by justifying the consistency between the content-matched identity perceptual watermark and the recovered robust watermark, without requiring the ground-truth. Moreover, source tracing can be accomplished based on the identity semantics that the recovered watermark carries. Extensive experiments demonstrate state-of-the-art detection and source tracing performance against Deepfake face swapping with promising watermark robustness for both cross-dataset and cross-manipulation settings.
\end{abstract}


\section{Introduction}
\label{sec:intro}

\IEEEPARstart{T}{he} prevalent Deepfake face swapping technique has received widespread attention with the development of deep generative models. By swapping the facial identity from the source image onto the target one, Deepfake face swapping can bring convenience to people’s lives and benefit society from the perspective of human entertainment such as film-making. Nevertheless, misusing the technique has brought privacy issues to various groups of victims including celebrities, politicians, and even every human being~\cite{DeepfakeSurvey[22]}. Therefore, it is necessary to prevent malicious attacks while maintaining positive usages of Deepfake face swapping.

Most existing countermoves~\cite{FFPP[11],XRay[12],MAT[13],NoiseDF[14],IIE[28],FakeDetect[41],AltFreeze[42],FullyUnsupervised[62]} regarding Deepfake face swapping follow the pipeline of devising a deep detection model that passively investigates and analyzes the underlying synthetic artifacts to distinguish the fake images. While accomplishing progresses in the early stages, passive detection suffers the generalization challenge such that obvious performance damping occurs when attempting to detect fake materials produced by unknown synthetic algorithms~\cite{Generalization[45]}. Moreover, following rapid improvements in deep generative algorithms, it has become burdensome to locate perceivable manipulation traces within high-quality synthetic images. To overcome this issue, several recent studies consider executing proactive defense against synthetic manipulations. In other words, the goal is to protect the original images by inserting imperceptible signals in advance of posting to the public and being manipulated by Deepfake unexpectedly.

\IEEEpubidadjcol

The proactive approaches against Deepfake face swapping can be categorized into two main strategies, namely, distorting~\cite{CMUA[1],AntiForgery[2],Disrupter[3]} and watermarking~\cite{FaceGuard[8],FaceSigns[9],Waterlo[37],VisibleMark[60],BiFPro[52]}. The former learns a distortion that nullifies potential synthetic manipulations by inserting it into the original images, and the latter encodes watermarks into original images and determines the authenticity based on the recovered watermarks afterward. Both ways aim to proactively protect the original images without affecting the visual qualities. While demonstrating preliminary initiations, the existing approaches suffer several flaws. Firstly, in the current research stage, both the distorting and watermarking approaches lack generalization ability on unseen datasets when facing unknown synthetic manipulations. Secondly, although the semi-fragile watermarks can help determine real and fake based on their existence, they are prone to be unstable when facing unavoidable common image processing operations. Moreover, watermarks that are fragile to malicious synthetic manipulations cannot help conduct source tracing on the victim target images simultaneously. In particular, in an integral chain of evidence for the forensic investigation in the cybercrime cases related to Deepfake face swapping, besides correctly addressing the falsification, it is crucial to retrieve intelligence information of the source materials~\cite{Sourcing[46],Sourcing[47],SepMark[10]}. In other words, tracing the original target images along with the detection task can help fulfill the evidence integrality. On the other hand, although few robust watermarking studies have been attempted, they generally determine real and fake relying on the ground-truths because the watermarks themselves are semantically void, making the proactive detection pipeline incomplete when the ground-truths are unavailable in real-life cases. Lastly, while preventing privacy issues, directly adding distortions not only unavoidably degrades the visual qualities of the images~\cite{Defeating[23]}, but also unfavorably disables benign adoptions of Deepfake face swapping.

To address the aforementioned issues, in this study, we propose a novel robust identity perceptual watermarking framework that promotes proactive defense against Deepfake face swapping. In general, we establish semantically meaningful identity perceptual watermarks that are robust and invisible based on image contents and embed them into original images for detection and source tracing purposes simultaneously, while allowing benign usages of face swapping algorithms. First, based on the identities of facial images, we construct securely protected identity perceptual watermarks by introducing a chaotic encryption system. Then, we devise an end-to-end robust watermarking workflow that consists of an encoder-decoder architecture to be trained with common and face swapping manipulation pools. The model is pre-trained with the common image processing operations to accomplish robustness with good visual quality when facing benign manipulations. Thereafter, we tune it with a single representative face swapping algorithm to achieve the objectives. Relying on the promising watermark recovery performance, the authenticity of a candidate image is determined by evaluating the consistency between its content-matched identity perceptual watermark and the robustly recovered watermark, without requiring ground-truth watermarks. The source tracing is simultaneously achieved with respect to the identity semantics carried by the recovered watermark. Extensive experiments prove the generalization ability of our proposed approach under cross-dataset and cross-manipulation settings, with average watermark recovery accuracies strongly outperforming the contrastive methods. Further, the outstanding detection performance against four GAN-based and four diffusion-based face swapping algorithms also consistently outperforms the state-of-the-art passive Deepfake detectors. Meanwhile, the average top-5, top-3, and top-1 source tracing accuracies of $98.10\%$, $97.72\%$, and $96.56\%$ against the eight face swapping models demonstrate the reliable ability of our proposed method to trace back to the original target facial identities. 

Our contributions can be summarized as follows:
\begin{itemize}
\item We propose a novel idea of identity perceptual watermarks based on facial image contents to proactively defend against Deepfake face swapping regarding content-watermark consistencies, without requiring ground-truths. A chaotic encryption system is collaboratively devised to ensure watermark confidentiality. 
\item We present an encoder-decoder workflow to invisibly embed watermarks into facial images and robustly recover them. Upon maintaining watermark robustness, our approach can concurrently achieve detection and source tracing for Deepfake face swapping by embedding and decoding a single robust watermark.  
\item Experiments qualitatively and quantitatively demonstrate the promising watermark recovery accuracy, Deepfake face swapping detection performance, and source tracing ability of our proposed approach, along with outstanding generalization ability against both GAN-based and diffusion-based generative models, outperforming state-of-the-art algorithms.
\end{itemize}

\section{Related Work}
\label{sec:related_work}

Existing forensic investigations on Deepfake face swapping mostly rely on devising passive detectors~\cite{FFPP[11],XRay[12],MAT[13],IIE[28],DeepfakeDiscrepancy[61],FakeDetect[41],AltFreeze[42]} that analyze traces of suspected images in various feature domains. In view of the rising challenge prompted by the improving synthetic quality due to well-developed generative models, instead of searching for synthetic artifacts, proactive defense aims to protect original images by inserting imperceptible signals before potential manipulations. In other words, while malicious face swapping happens after the target image is publicly posted (e.g., on Twitter or Instagram), proactive actions are usually performed in advance in two ways: \textit{distortions} and \textit{watermarks}.

\subsection{Proactive Distortions}

Proactive distortions~\cite{Disrupting[32],InitDefense[34],Nullify[49],Disrupting_WACVW[35],TAFIM[36]} are usually inserted into pristine images to nullify the synthetic models following $\tilde{x} = x + \eta$, where $x$ and $\tilde{x}$ are the pristine images before and after the learned invisible distortion $\eta$ is added, respectively. An early work~\cite{segalis2020ogan[63]} introduces the oscillating GAN (OGAN) attack, which successfully creates visible oscillations in synthetic facial images. Recently, Anti-Forgery~\cite{AntiForgery[2]} exploits the vulnerability of GAN-based models and supervisely learns a particular distortion in the lab color space for a specific pristine image regarding each synthetic algorithm. CMUA~\cite{CMUA[1]} presents a two-level perturbation fusion strategy and heuristically learns a universal distortion that can disrupt multiple synthetic algorithms. TAFIM~\cite{TAFIM[36]} introduces image-specific perturbations with better robustness via attention-based fusion and refinement. Besides adding invisible perturbations that could heavily distort the synthetic results, Wang \textit{et al.}~\cite{Disrupter[3]} further promoted a detector that distinguishes the distorted images with high accuracy since machines do not share similar perspectives as human eyes.

\subsection{Proactive Watermarks}

Digital watermarks~\cite{Watermark[43]} are originally used for copyright protection and authentication~\cite{Ownership[29],Ownership[30],Ownership[31]}, and they are thus designed to have promising robustness. Ever since the occurrence of the first deep neural network (DNN) based end-to-end trainable watermarking framework, HiDDeN~\cite{Hidden[4]}, various follow-up algorithms~\cite{MBRS[5],CIN[6],StegaStamp[7],PIMoG[40],ARWGAN[53]} emerge based on the encoder-decoder architecture for robust watermarking. The encoder reconstructs a visually identical image by inserting a secret message into the original one, and the decoder takes charge of recovering the encoded message from the watermarked image. Particularly, MBRS~\cite{MBRS[5]} proposes a novel training strategy within mini-batches to enhance the JPEG robustness of watermark embedding. To simultaneously improve watermark imperceptibility and robustness against unknown image noises, CIN~\cite{CIN[6]} combines the invertible and non-invertible mechanisms in the watermark embedding and recovery pipelines. ARWGAN~\cite{ARWGAN[53]} enforces more abundant features from images for robust watermarking such that the image visual quality and watermark recovery accuracy can be concurrently boosted. 

\begin{figure*}[t]
  \centering
   \includegraphics[width=\linewidth]{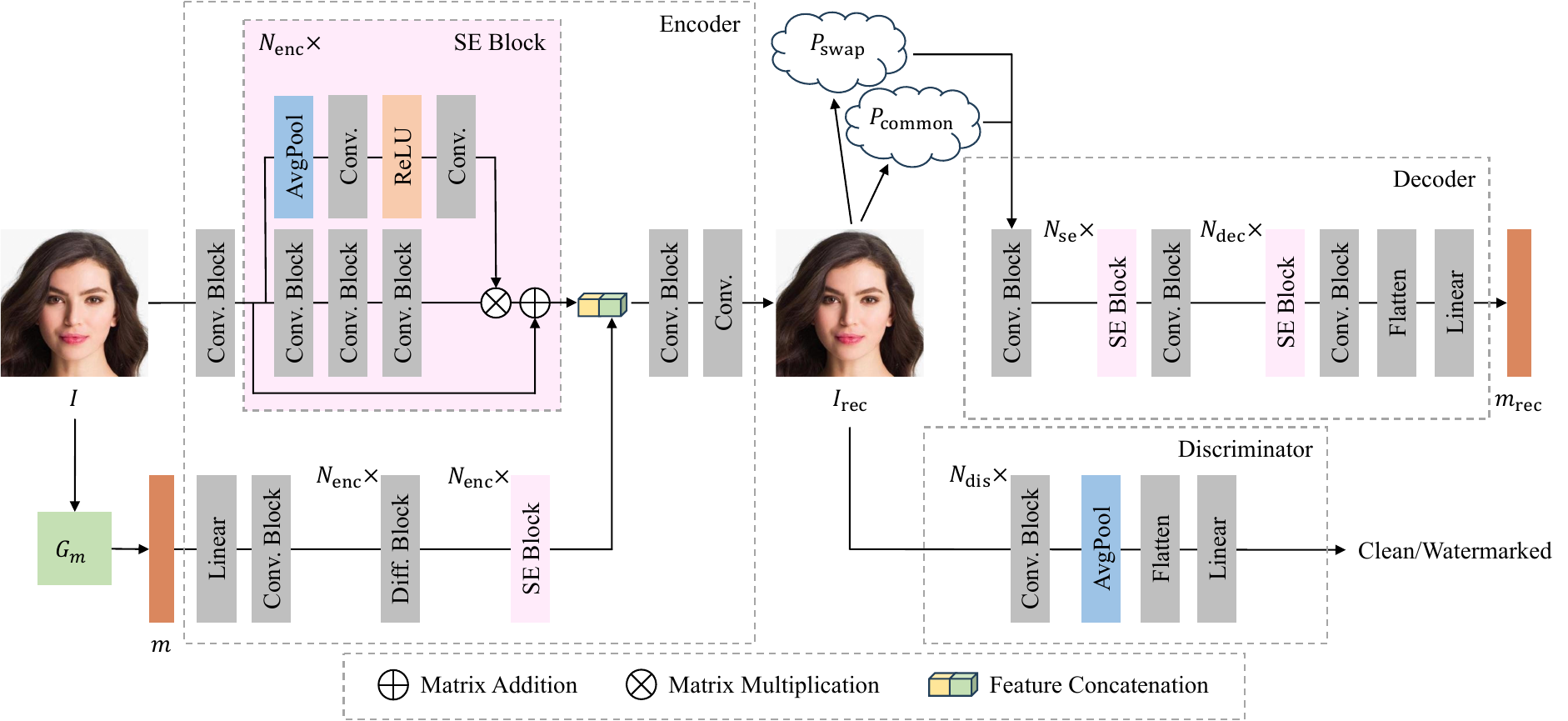}
   \caption{Framework of the proposed approach. An original clean image $I$ is first passed to generator $G_m$ to generate the corresponding identity perceptual watermark $m$. Then, $I$ and $m$ are fed to the encoder for watermark embedding, which derives $I_\textrm{rec}$. Manipulation pools $P_\textrm{common}$ and $P_\textrm{swap}$ containing common and Deepfake manipulations help enhance watermark robustness during supervision. Besides, a discriminator helps maintain the visual quality. In the end, the watermark $m_\textrm{rec}$ is recovered by passing the manipulated image to the decoder.}
   \label{fig:framework}
\end{figure*}

Regarding the relatively novel term of proactive defense in digital forensics, few approaches have been attempted with watermarks~\cite{RootDeepfake[33],SepMark[10],Zhao2024Proactive[65],Zhao2023WACV_Proactive[66],meng2022traceable[73]}. Specifically, Yu \textit{et al.}~\cite{RootDeepfake[33],ScalableGAN[48]} assigned particular fingerprints to images via robust watermarking regarding generative models for the attribution purpose. Wang \textit{et al.}~\cite{Wang2021FakeTagger[64]} raised the importance of Deepfake provenance tracking and designed FakeTagger to perform image tagging via a novel channel-coding injection operation. Different from the robust watermarking studies, FaceGuard~\cite{FaceGuard[8]} and FaceSigns~\cite{FaceSigns[9]} propose semi-fragile watermarks such that they stay robust to benign image post-processing operations but become fragile when facing malicious face manipulations. The authors claimed to distinguish fake materials if no watermark is detected in candidate images. Meanwhile, Zhao \textit{et al.}~\cite{Zhao2024Proactive[65]} introduced a semi-fragile watermark that performs promisingly against the classical attacks including copy-move, inpainting, and splicing. Later, IMD~\cite{asnani2022proactive[70]} and MaLP~\cite{asnani2023pro_loc[69]} are proposed to detect Deepfake faces by inserting pre-learned semi-fragile templates and computing the cosine similarity between extracted and ground-truth templates, where MaLP is further expanded to all image contents and perform localization based on IMD. Recently, a separable watermarking framework, SepMark~\cite{SepMark[10]}, is conducted with two watermark decoders, a Tracer and a Detector, where the former recovers robust watermarks for source tracing, and the latter extracts semi-fragile watermarks to determine real and fake. Similarly, FakeTracer~\cite{sun2022faketracer[68],sun2024faketracer[67]} introduces robust and semi-fragile traces, STrace and ETrace, to detect Deepfake face swapping according to the co-existence of the traces. Differently, FakeTracer requires the traces to be implanted into images before the training phase of face swapping models such that the generative models learn to maintain only STrace during inference. To the best of our knowledge, compared to the abundance in defending against facial attribute editing, robust watermarking has been barely attempted to detect Deepfake face swapping adaptively. Moreover, performing detection and source tracing simultaneously without ground-truths via a single watermark remains largely unexplored in the current research domain. 

\section{Methodology}

\subsection{Overview}

In this paper, we propose a novel robust identity perceptual watermarking framework to adaptively embed imperceptible robust watermarks into images and proactively achieve detection and source tracing against Deepfake face swapping by precisely recovering the encoded watermarks. The framework of our approach is demonstrated in Fig.~\ref{fig:framework}. First, we devise a watermark generator $G_m$ to securely prepare the binary watermarks by assigning identity perceptual semantics with respect to the image contents. Then, the encoder is fed with the clean image $I$ and the corresponding identity perceptual watermark $m$ for encoding. To guarantee robustness, we apply a common manipulation pool $P_\textrm{common}$ and a face swapping manipulation pool $P_\textrm{swap}$ as adversaries to the watermarked image $I_\textrm{rec}$ before recovering the watermark $m_\textrm{rec}$ via the decoder. Meanwhile, a discriminator that distinguishes the watermarked and clean images is adopted in the training stage to enhance image visual quality. By assigning constraints as discussed in Section~\ref{sec:loss_func}, the framework is trained to satisfy our objectives of watermark robustness. 

In practice, for an original image $I$ with facial identity $t$, the identity perceptual watermark $m$ is produced by $G_m$ and embedded into $I$ to generate the watermarked image $I_\textrm{rec}$ via the encoder. Upon malicious face swapping manipulations, a synthetic facial image is derived by swapping a different facial identity onto $I_\textrm{rec}$. By utilizing the decoder, the watermark $m_\textrm{rec}$, which is faithfully consistent with $m$, can be robustly recovered from the synthetic facial image. Then, the identity perceptual watermark of the synthetic facial image can be obtained via $G_m$. Since it contains a different facial identity, its corresponding watermark is different from the robustly recovered watermark $m_\textrm{rec}$, and demonstrating the inconsistency between the two watermarks accomplishes Deepfake detection on the synthetic facial image. On the other hand, if the watermark-protected image $I_\textrm{rec}$ does not undergo face swapping but certain common image operations, the identity perceptual watermark obtained via $G_m$ from the suspect image is identical to $m_{\textrm{rec}}$, which is then classified as real content. Moreover, as $m_\textrm{rec}$ is robustly recovered, the original image $I$ with identity perceptual watermark $m$ can be traced based on watermark semantics by inspecting the matching between identities and watermarks.

\subsection{Identity Perceptual Watermarks}
\label{sec:id_watermarks}

Unlike early approaches that enforce random watermark initialization, we derive the watermark for each facial image regarding the corresponding identity semantics, which then allows the absence of ground-truths in proactive Deepfake detection. In particular, we propose an identity perceptual watermark generator $G_m$ that derives a securely encrypted binary watermark $m$ of length $l$ based on the facial identity of the image $I$. Specifically, inspired by hashing algorithms~\cite{Perceptual[50],Hash[51]}, the pipeline of $G_m$ follows two critical characteristics: \textit{collision free} and \textit{one-way computational}.

\noindent\textbf{Collision Free.} Since the goal is to detect Deepfake face swapping by addressing inconsistencies between the recovered and content-perceptual watermarks of candidate suspect images, it is necessary to ensure that no two identical watermarks belong to different facial identities. Face recognition, a task to identify different facial identities by producing distinguishable embeddings, has been proved the ability on various large facial datasets with reliable models. In this study, impressed by the outstanding achievements, we adopt well-performed face recognition algorithms that retrieve identity embeddings in the form of high-dimensional vectors. To match up the required watermark length $l$, we first select a large dataset that contains sufficient unique identities and retrieve the identity embeddings $E$ with dimension $d_E$ accordingly. Then, we apply the principal component analysis (PCA)~\cite{PCA[24]} to $E$ and learn a transform rule $\phi$ that projects the identity embeddings of the images from dimension $d_E$ down to $l$ while optimally preserving the original correlations between facial identities with respect to the principal components. Thereafter, for the purpose of constructing binary watermarks, we conduct a unit-variance scaling on the projected embeddings $E_\textrm{pca}$ to the range between 0 and 1 by
\begin{equation}
E_\textrm{scale} = \frac{E_\textrm{pca} - \min(E_\textrm{pca})}{\max(E_\textrm{pca}) - \min(E_\textrm{pca})},
\label{eq:min_max_scale}
\end{equation}
where $E_\textrm{scale}$ denotes the scaled embeddings with respect to the standard deviation, which favorably preserves the distribution of $E_\textrm{pca}$. In the end, we set a cutoff value $0 \leq c \leq 1$ to round the entries of $E_\textrm{scale}$ to binary values accordingly, and obtain the identity perceptual watermarks $E_\textrm{binary}$. Besides, a watermark collision check is arranged in Section~\ref{sec:collision_check} to validate the collision-free characteristic. 

\noindent\textbf{One-Way Computational.} Since we concurrently conduct detection and source tracing based solely on robust watermarks, full access to the watermarks is provided to the users. Therefore, an adversary with sufficient knowledge of our work may aim to attack accordingly, such as replacing the encoded watermark with a content-matched one. To prevent hostile attacks~\cite{HostilAttack[38],HostilAttack[39]}, we enforce the watermark to be one-way computational, that is, it is insufficient to recover the watermark construction pipeline by merely acquiring the watermark from the image. Specifically, besides keeping $\phi$ encapsulated, we design a chaotic encryption~\cite{ChaoticEnc[44]} system on $E_\textrm{binary}$ using a logistic map
\begin{equation}
x_{i+1} = rx_i(1 - x_i),
\label{eq:logistic_map}
\end{equation}
where constants $x_0$ at $i=0$ and $r$ are the initial condition and control parameter, respectively. $x_i$ denotes the $i$-th value in the chaotic map and $0 <= i <l$. The values in $X=\{x_1, x_2, ..., x_l\}$ are derived accordingly, and a binary encryption key $K=\{k_1, k_2, ..., k_l\}$ is computed via
\begin{equation}
k_i = \lfloor x_i p^q \bmod{2} \rfloor,
\label{eq:key_encrypt}
\end{equation}
where $p$ and $q$ are pre-defined prime constants. Finally, the encrypted identity perceptual watermark $m$ for image $I$ is derived by a logical exclusive OR (XOR) operation between $K$ and the unencrypted watermark $e_\textrm{binary} \in E_\textrm{binary}$. Relying on the sensitive dependence on the initial condition and chaotic map nonlinearity characteristics, the final watermarks leveraged in the watermarking framework are unpredictable and nonreversible unless having access to Eqn.($\ref{eq:logistic_map}$) and Eqn.($\ref{eq:key_encrypt}$) including the critical constant coefficients. Specifically, the entry values in $X$ vary drastically with respect to the values of $x_0$ and $r$, and the watermarks are thus securely protected. On the other hand, with authority on $K$, performing an extra XOR operation on $m$ using $K$ can easily decrypt the watermarks back to the ones in $E_\textrm{scale}$ to trace back for the identity information.

\subsection{Watermark Encoding}

For an input image $I$, we embed the watermark $m$ using an encoder that is mainly based on convolutional neural networks (CNN) with squeeze-and-excitation networks (SENet)~\cite{SENet[25]} and diffusion blocks. Specifically, image and watermark features are separately analyzed and then concatenated to reconstruct the watermarked image $I_\textrm{rec}$. 

\noindent\textbf{Image Features.} In this paper, we preserve the width and height of $I$ in the propagation process to avoid information loss in image reconstruction. We first feed $I$ to a convolutional block (denoted as Conv. Block) that contains a CNN layer, a batch normalization layer, and a ReLU activation function in a sequential order to exaggerate the number of feature channels. Squeeze-and-excitation network (SENet)~\cite{SENet[25]}, an on-the-shelf mechanism that improves channel interdependencies, can favorably perform feature recalibration on the input image. As depicted in Fig.~\ref{fig:framework}, $N_\textrm{enc}$ repeated SENets (denoted as SE Block) are applied to seek proper channels and elements within the feature map for watermark encoding without jeopardizing the visual quality. 

\noindent\textbf{Watermark Features.} As for the securely encryptedd binary identity perceptual watermark $m$ of length $l$, we first increase the length linearly via a fully connected layer followed by dimension expansion via shape rearrangement, and then leverage a convolutional block to elaborate the channel perspective. After that, we devise $N_\textrm{enc}$ consecutive diffusion blocks (denoted as Diff. Block) utilizing deconvolutions to smoothly diffuse the feature elements that represent $m$ to the same dimension as the width and height of $I$. Similarly, a series of $N_\textrm{enc}$ SE Blocks is arranged for further feature refinement for the preparation of watermark embedding. 

\noindent\textbf{Image Reconstruction.} Lastly, features of image $I$ and watermark $m$ are concatenated along the channel dimension, followed by a convolutional block and a vanilla CNN layer to upscale features and reconstruct the image $I_\textrm{rec}$ that is visually identical to $I$ but with $m$ invisibly embedded.

\subsection{Watermark Recovery}

\noindent\textbf{Manipulation Pools.} In real-life scenarios, images inevitably suffer degradation due to common post-processing operations and noises such as compression and blurring upon uploading and spreading on social networks. Therefore, to satisfy the objectives, we construct two manipulation pools, namely $P_\textrm{common}$ and $P_\textrm{swap}$, to supervise the proposed framework. The common manipulation pool $P_\textrm{common}$ contains the benign post-processing operations that are commonly seen in the real world. On the other hand, we include Deepfake face swapping models in the face swapping pool $P_\textrm{swap}$ to preserve watermark robustness when facing malicious attacks.

\noindent\textbf{Decoder.} We feed the post-manipulation image to the decoder for watermark recovery. The decoder follows a reverse pipeline of the encoder. In particular, after extracting features back to a large number of channels, we apply $N_\textrm{se}$ SE Blocks to gradually expand the number of channels and seek determinant elements that hide key watermark information. Then, $N_\textrm{dec}$ SE Blocks without channel modification are performed after narrowing down the channel number. In the end, a convolutional block, a flattening operation, and a linear projection are sequentially applied to recover the watermark $m_\textrm{rec}$, which is faithfully identical to the original watermark $m$.

\begin{figure*}
\centering
   \includegraphics[width=\textwidth]{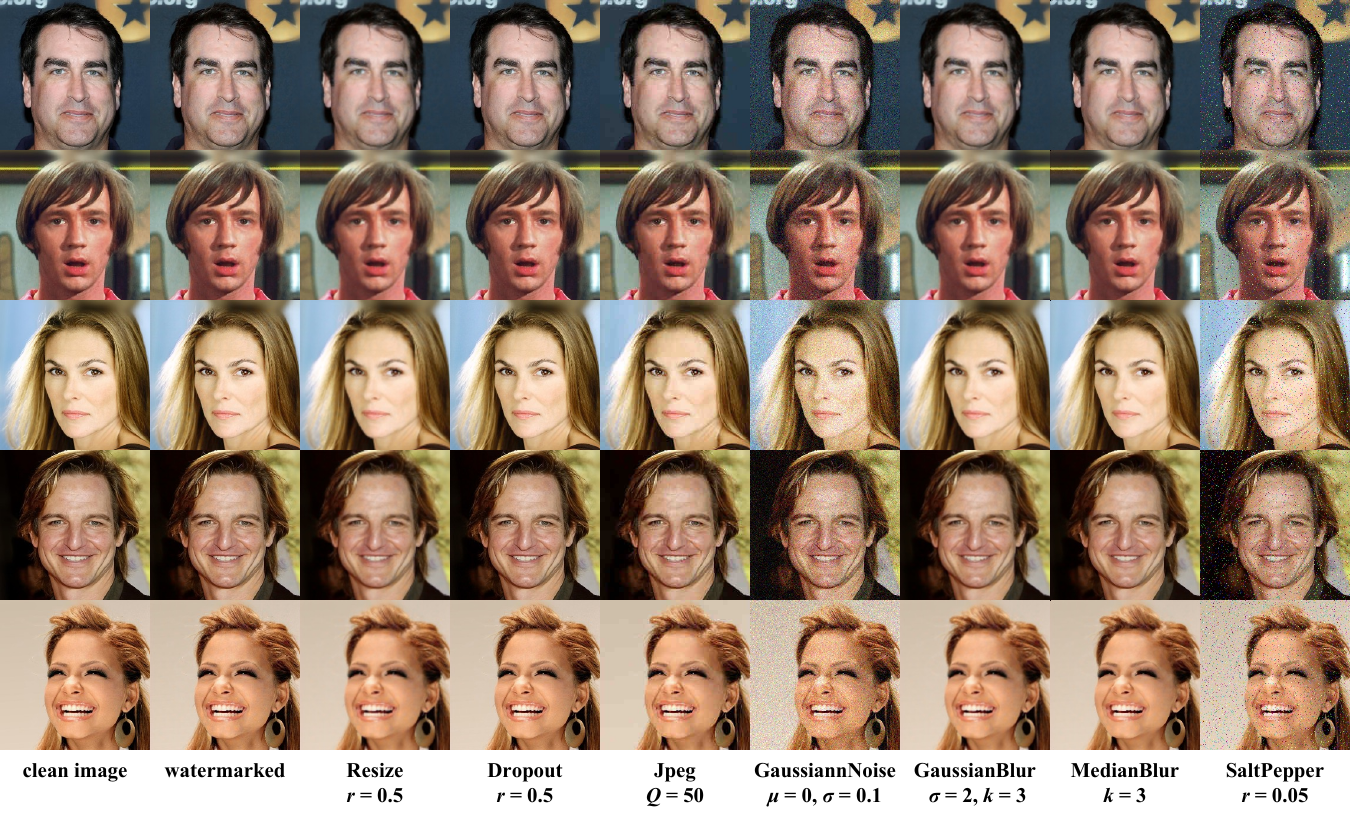}
   \caption{Visual effects of common image manipulations on the watermarked images. The first two columns refer to the raw and watermarked images, and each of the rest rows displays the visual effects of a manipulation algorithm on the watermarked images. }
   \label{fig:common_exhibition}
\end{figure*}

\subsection{Objective Functions}
\label{sec:loss_func}

During the training stage, we employ a total of 4 loss functions for guidance, namely, \textit{reconstruction loss}, \textit{recovery loss}, \textit{adversarial loss}, and \textit{generative loss}.

\noindent \textbf{Reconstruction Loss.} We embed the watermarks into images and pursue the uttermost similarity between the original and watermarked images by assigning a pixel-wise $L_2$ constraint on the encoder following
\begin{equation}
L_\textrm{enc} = \lVert \textrm{Enc}(\theta_\textrm{enc}, I, m) - I \rVert_2,
\label{eq:enc_loss}
\end{equation}
where $I$ denotes the clean image before watermarking, $m$ represents the watermark, and $\theta_\textrm{enc}$ encloses the parameters of the encoder $\textrm{Enc}(\cdot)$.

\noindent \textbf{Recovery Loss.} Similarly, a bit-wise $L_2$ constraint is applied on the decoder for watermark recovery following
\begin{equation}
L_\textrm{dec} = \lVert \textrm{Dec}(\theta_\textrm{dec}, \textrm{Enc}(\theta_\textrm{enc}, I, m)) - m \rVert_2,
\label{eq:dec_loss}
\end{equation}
where $\theta_\textrm{dec}$ refers to the parameters of the decoder $\textrm{Dec}(\cdot)$. The encoder is updated alongside.

\noindent \textbf{Adversarial Loss.} Following the convention of robust watermarking~\cite{Hidden[4],MBRS[5],CIN[6]}, the discriminator $D$ with parameters $\theta_d$ is applied to improve the visual qualities of the images following 
\begin{equation}
L_\textrm{adv} = -\mathbb{E} (\log (D(\theta_d, I))) + \mathbb{E} (\log (1 - D(\theta_d, \textrm{Enc}(\theta_\textrm{enc}, I, m)))).
\label{eq:adv_loss}
\end{equation}
The discriminator is implemented as a binary classifier with $N_\textrm{dis}$ convolutional blocks, and it adversarially learns to identify clean and watermarked images for each pair of $I$ and $I_\textrm{rec}$ during model supervision.

\noindent \textbf{Generative Loss.} When training with the face swapping manipulation pool $P_\textrm{swap}$, we arrange an extra objective using an $L_2$ constraint to make sure the embedded watermarks do not affect the synthetic results following
\begin{equation}
L_\textrm{gen} = \lVert G(I, I_s) - G(\textrm{Enc}(\theta_\textrm{enc}, I, m), I_s) \rVert_2,
\label{eq:gen_loss}
\end{equation}
where $I_s$ is the image that provides identity information for the face swapping model $G(\cdot)$.

\noindent \textbf{Overall Objective Function.} Coefficients $\lambda_\textrm{enc}$, $\lambda_\textrm{dec}$, $\lambda_\textrm{adv}$, and $\lambda_\textrm{gen}$ are assigned to compute the weighted sum in the overall objectives $L_\textrm{common}$ and $L_\textrm{swap}$ for training with $P_\textrm{common}$ and $P_\textrm{swap}$ following
\begin{equation}
L_\textrm{common} = \lambda_\textrm{enc}L_\textrm{enc} + \lambda_\textrm{dec}L_\textrm{dec} + \lambda_\textrm{adv}L_\textrm{adv},
\label{eq:total_loss_common}
\end{equation}
and
\begin{equation}
L_\textrm{swap} = L_\textrm{common} + \lambda_\textrm{gen}L_\textrm{gen}.
\label{eq:total_loss_swap}
\end{equation}

\section{Experiment}


\subsection{Implementation Details}
\label{sec:impl_detail}

\noindent\textbf{Datasets.} Experiments are conducted on the high-quality CelebA-HQ dataset~\cite{CelebAHQ[15]} with 30,000 image samples and 6,217 unique identities following the official split~\cite{CelebA[16]} for training, validation, and testing. To further validate the generalization ability, we tested the trained model on the Labeled Faces in the Wild (LFW) dataset~\cite{LFW[17]} with 5,749 unique identities. 

\noindent\textbf{Manipulation Pools.} $P_\textrm{common}$ consists of the following manipulations for both training and testing: Dropout, Resize, Jpeg Compression, Gaussian Noise, Salt and Pepper, Gaussian Blur, and Median Blur. Contrarily, $P_\textrm{swap}$ includes SimSwap~\cite{SimSwap[18]} solely in the training phase and adopts SimSwap, InfoSwap~\cite{InfoSwap[19]}, UniFace~\cite{UniFace[20]}, E4S~\cite{e4s[54]}, DiffSwap~\cite{DiffSwap[55]}, FuseAnyPart~\cite{FuseAnyPart[76]}, Face-Adapter~\cite{Face-Adapter[77]}, and REFace~\cite{REFace[78]} with state-of-the-art performance for cross-manipulation evaluation in the testing phase. Since there are various common manipulations to be considered and they only bring noises without modifying the facial contents, we first pre-trained the watermarking model using $P_\textrm{common}$ for sufficient iterations. Henceforth, we slowly fine-tuned it with the help of $P_\textrm{swap}$ by introducing random images that provide different facial identities to be swapped to $I_\textrm{rec}$.

\noindent\textbf{Parameters.} Upon watermark generation, CelebA-HQ is adopted to derive the transform rule $\phi$ with cutoff value $c = 0.5$, and we set $x_0 = 0.1$, $r = 3.93$, $p = 5$, and $q = 11$ for the constant coefficients in Eqn.($\ref{eq:logistic_map}$) and Eqn.($\ref{eq:key_encrypt}$). We conducted watermark encoding with lengths $64$ and $128$ for images at $128 \times 128$ and $256 \times 256$ resolutions, respectively, and we set $N_\textrm{enc}$ to $3$ and $4$, $N_\textrm{dec}$ to $1$ and $2$, $N_\textrm{se}$ to $4$ and $5$, and $N_\textrm{dis}$ to $3$ and $4$ for the two resolutions. In this study, we trained the model in two steps. The proposed approach is firstly trained with $P_\textrm{common}$ by setting $\lambda_\textrm{enc} = 1$, $\lambda_\textrm{dec} = 10$, and $\lambda_\textrm{adv} = 0.01$ with a learning rate of $1e-3$ to guarantee the promising watermark recovery performance regarding common manipulations. Then, the trained model is further tuned on SimSwap~\cite{SimSwap[18]} by setting $\lambda_\textrm{enc} = 10$, $\lambda_\textrm{dec} = 1$, $\lambda_\textrm{adv} = 0.01$, and $\lambda_\textrm{gen} = 5$ with a lower learning rate of $1e-5$. This process mildly adjusts the framework to stay robust when facing both common and face swapping manipulations. Adam optimizers~\cite{Adam[26]} are adopted through the entire training pipeline for each model component. Our model is implemented using PyTorch on $4$ Tesla V100 GPUs with a batch size of $32$.

\subsection{Experiment on CelebA-HQ}
\label{sec:exp_celeba}

In this section, experiments are first conducted on the testing set of CelebA-HQ. Since no existing work has performed detection and source tracing on face swapping images using a single watermark to the best of our knowledge, we selected the robust watermarking frameworks (HiDDeN~\cite{Hidden[4]}, MBRS~\cite{MBRS[5]}, and CIN~\cite{CIN[6]}), proactive robust Deepfake watermarking frameworks (RDA~\cite{RootDeepfake[33]}, ARWGAN~\cite{ARWGAN[53]}, SepMark\footnote{SepMark contains a robust and a semi-fragile watermark. Since the watermark robustness is evaluated, we picked the robust watermark of SepMark for comparison in experiments. }~\cite{SepMark[10]}, WEvade~\cite{jiang2023evading[74]}, and InvisMark~\cite{InvisMark[75]}), proactive semi-fragile Deepfake watermarking frameworks (IMD~\cite{asnani2022proactive[70]}, MaLP~\cite{asnani2023pro_loc[69]}, and FaceSigns~\cite{FaceSigns[9]}), and proactive Deepfake distortion frameworks (AntiForgery~\cite{AntiForgery[2]} and CMUA~\cite{CMUA[1]}) for comparison. For fairness, the publicly available trained model weights that correspond to optimal performance are directly adopted. 

\begin{figure*}[t!]
  \centering
   \includegraphics[width=\linewidth]{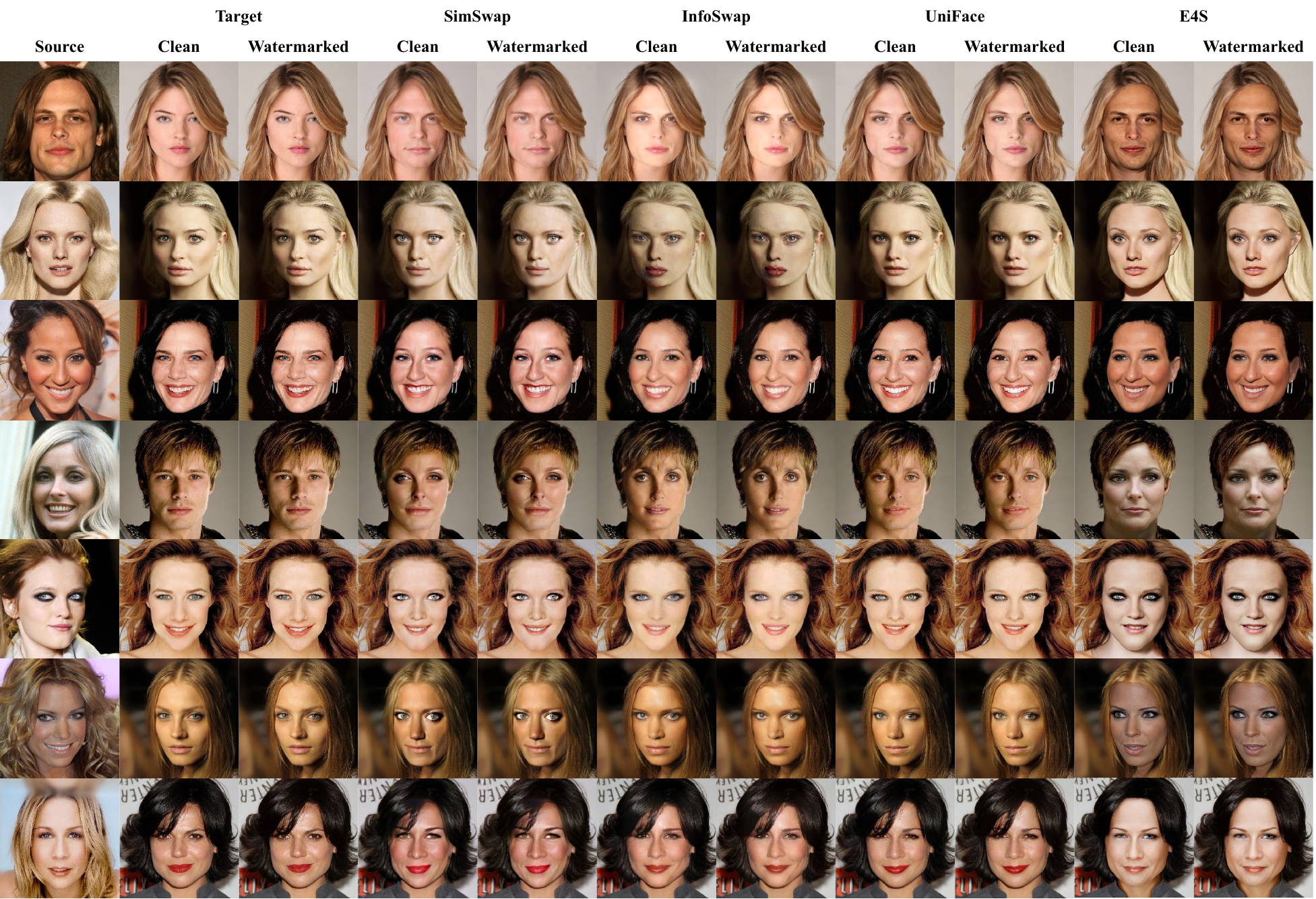}
   \caption{Visual effects of Deepfake face swapping image manipulations on clean and watermarked images. The first column refers to the source images that provide facial identity for face swapping. Columns 2 and 3 present the raw and watermarked target images. Every two columns of the remaining exhibit the face swapping visualizations on the raw and watermarked target images. }
   \label{fig:faceswap_exhibition}
\end{figure*}

\begin{table}[t!]
  \centering
  \caption{Quantitative visual quality evaluation of the watermarked images on CelebA-HQ. Information includes model name, image resolution, watermark length, PSNR, and SSIM. The best performance for each resolution is in \textbf{bold} and the second best performance is marked with \underline{underscore}.}
  \begin{tabular}{@{}lcccc@{}}
    \toprule
    Model & Resolution & Length & PSNR$\uparrow$ & SSIM$\uparrow$ \\
    \midrule
    HiDDeN~\cite{Hidden[4]} & $128 \times 128$ & 30 & 33.26 & 0.888 \\
    MBRS~\cite{MBRS[5]} & $128 \times 128$ & 30 & 33.01 & 0.775 \\
    RDA~\cite{RootDeepfake[33]} & $128 \times 128$ & 100 & \underline{43.93} & \underline{0.975} \\
    CIN~\cite{CIN[6]} & $128 \times 128$ & 30 & 43.37 & 0.967 \\
    IMD~\cite{asnani2022proactive[70]} & $128 \times 128$ & -- & 34.84 & 0.841 \\
    MaLP~\cite{asnani2023pro_loc[69]} & $128 \times 128$ & -- & 34.87 & 0.844 \\
    ARWGAN~\cite{ARWGAN[53]} & $128 \times 128$ & 30 & 39.58 & 0.919 \\
    SepMark~\cite{SepMark[10]} & $128 \times 128$ & 30 & 38.51 & 0.959 \\
    WEvade~\cite{jiang2023evading[74]} & $128 \times 128$ & 30 & 34.56 & 0.800  \\
    Ours & $128 \times 128$ & 64 & \textbf{47.39} & \textbf{0.993} \\
    \midrule
    AntiForgery~\cite{AntiForgery[2]} & $256 \times 256$ & -- & 35.62 & 0.953 \\
    CMUA~\cite{CMUA[1]} & $256 \times 256$ & -- & 38.64 & 0.857 \\
    MBRS~\cite{MBRS[5]} & $256 \times 256$ & 256 & 44.14 & 0.969 \\
    FaceSigns~\cite{FaceSigns[9]} & $256 \times 256$ & 128 & 36.99 & 0.889 \\
    SepMark~\cite{SepMark[10]} & $256 \times 256$ & 128 & 38.56 & 0.933 \\
    InvisMark~\cite{InvisMark[75]} & $256 \times 256$ & 100 & \underline{44.20} & \textbf{0.997} \\
    Ours & $256 \times 256$ & 128 & \textbf{45.38} & \underline{0.994} \\
    \bottomrule
  \end{tabular}
  \label{tab:vis_quality_celeba}
\end{table}

\begin{table*}[t!]
  \centering
  \caption{Quantitative comparison on CelebA-HQ at the $128 \times 128$ resolution regarding the bit-wise recovery accuracy of the watermarks under common manipulations and malicious face swapping manipulations. The best performance for each resolution is in \textbf{bold} and the second best performance is marked with \underline{underscore}.}
  \resizebox{\textwidth}{!}{
  \begin{threeparttable}
  \begin{tabular}{@{}l|cccccccccc@{}}
    \toprule
    Manipulations & HiDDeN~\cite{Hidden[4]} & MBRS~\cite{MBRS[5]} & RDA~\cite{RootDeepfake[33]} & CIN~\cite{CIN[6]} & IMD$^\dagger$~\cite{asnani2022proactive[70]} & MaLP$^\dagger$~\cite{asnani2023pro_loc[69]} & ARWGAN~\cite{ARWGAN[53]} & SepMark~\cite{SepMark[10]} & WEvade~\cite{jiang2023evading[74]} & Ours \\
    \midrule
    Dropout & 82.44\% & 99.99\% & 94.76\% & 99.99\% & 0.5573 & 0.4532 & 97.25\% & 99.99\% & 78.02\% & 99.99\% \\
    Resize & 82.01\% & 99.99\% & 99.94\% & 99.17\% & 0.4092 & 0.4562 & 93.73\% & 99.76\% & 69.67\% & 99.99\% \\
    Jpeg & 67.84\% & 99.49\% & 66.85\% & 96.86\% & 0.5084 & 0.1379 & 57.42\% & 98.14\% & 66.69\% & 99.48\% \\
    GaussianNoise & 51.36\% & 99.60\% & 60.18\% & 86.00\% & 0.4497 & 0.4458 & 53.45\% & 59.31\% & 64.87\% & 98.63\% \\
    SaltPepper & 52.30\% & 99.37\% & 66.62\% & 93.95\% & 0.4422 & 0.2804 & 64.55\% & 99.90\% & 55.79\% & 95.54\% \\
    GaussianBlur & 73.04\% & 99.99\% & 99.95\% & 99.99\% & 0.3609 & 0.4581 & 85.22\% & 99.30\% & 65.58\% & 99.99\% \\
    MedBlur & 82.72\% & 99.99\% & 99.98\% & 97.03\% & 0.4120 & 0.4544 & 96.66\% & 99.98\% & 68.12\% & 99.99\% \\
    \midrule
    Average & 70.24\% & \textbf{99.78\%} & 84.04\% & 96.14\% & 0.4485 & 0.3837 & 78.33\% & 93.77\% & 66.96\% & \underline{99.09\%} \\
    \midrule
    SimSwap~\cite{SimSwap[18]} & 50.02\% & 49.98\% & 50.00\% & 50.28\% & 0.0035 & 0.7228 & 52.06\% & 59.02\% & 59.25\% & 99.99\% \\
    InfoSwap~\cite{InfoSwap[19]} & 50.07\% & 50.82\% & 50.01\% & 50.60\% & 0.5261 & 0.1489 & 47.94\% & 76.62\% & 66.51\% & 99.31\% \\
    UniFace~\cite{UniFace[20]} & 54.98\% & 50.22\% & 71.15\% & 46.01\% & 0.4093 & 0.4467 & 59.30\% & 66.13\% & 51.50\% & 99.57\% \\
    E4S~\cite{e4s[54]} & 49.19\% & 50.07\% & 63.03\% & 50.55\% & 0.0071 & 0.6408 & 49.81\% & 79.72\% & 74.06\% & 93.19\% \\
    DiffSwap~\cite{DiffSwap[55]} & 40.53\% & 43.48\% & 61.94\% & 47.15\% & 0.5298 & 0.4935 & 58.30\% & 72.73\% & 78.77\% & 80.10\% \\
    FuseAnyPart~\cite{FuseAnyPart[76]} & 73.57\% & 88.95\% & 48.25\% & 66.88\% & 0.4891 & 0.6010 & 69.58\% & 90.73\% & 80.56\% & 91.00\% \\
    Face-Adapter~\cite{Face-Adapter[77]} & 73.12\% & 76.74\% & 50.42\% & 68.57\% & 0.3799 & 0.6109 & 71.84\% & 79.10\% & 75.67\% & 82.26\% \\
    REFace~\cite{REFace[78]} & 48.17\% & 89.03\% & 49.86\% & 88.56\% & 0.6206 & 0.6240 & 80.01\% & 90.52\% & 76.87\% & 94.61\% \\
    \midrule
    Average & 54.96\% & 62.41\% & 55.58\% & 58.58\% & 0.3707 & 0.5361 & 61.11\% & \underline{76.83\%} & 70.40\% & \textbf{92.50\%} \\
    \bottomrule
  \end{tabular}
  \begin{tablenotes}
    \small {
    \item{ $\dagger$ IMD and MaLP are specially evaluated via cosine similarity according to the original paper. Higher values denote better watermark robustness. }}
  \end{tablenotes}
  \end{threeparttable}
  }
  \label{tab:bit_accuracy_celebahq_128}
\end{table*}

\begin{table}[h!]
  \centering
  \caption{Quantitative comparison on CelebA-HQ at the $256 \times 256$ resolution regarding the bit-wise recovery accuracy of the watermarks under common manipulations and malicious face swapping manipulations. The best performance for each resolution is in \textbf{bold} and the second best performance is marked with \underline{underscore}.}
  \resizebox{\columnwidth}{!}{
  \begin{tabular}{@{}lcccccc@{}}
    \toprule
    Manipulations & MBRS~\cite{MBRS[5]} & FaceSigns~\cite{FaceSigns[9]} & SepMark~\cite{SepMark[10]} & InvisMark~\cite{InvisMark[75]} & Ours \\
    \midrule
    Dropout & 99.19\% & 55.36\% & 99.99\% & 99.59\% & 99.99\% \\
    Resize & 99.99\% & 98.63\% & 99.99\% & 99.99\% & 99.08\% \\
    Jpeg & 99.69\% & 82.64\% & 99.94\% & 99.91\% & 99.99\% \\
    GaussianNoise & 58.31\% & 53.61\% & 52.71\% & 99.79\% & 99.97\% \\
    SaltPepper & 65.33\% & 73.95\% & 99.99\% & 87.27\% & 97.54\% \\
    GaussianBlur & 72.05\% & 98.68\% & 99.92\% & 99.99\% & 97.93\% \\
    MedBlur & 98.06\% & 99.86\% & 99.99\% & 99.99\% & 99.01\% \\
    \midrule
    Average & 84.66\% & 80.39\% & 93.22\% & \underline{98.13\%} & \textbf{99.07\%} \\
    \midrule
    SimSwap~\cite{SimSwap[18]} & 50.00\% & 49.74\% & 54.07\% & 52.92\% & 99.98\% \\
    InfoSwap~\cite{InfoSwap[19]} & 50.71\% & 50.00\% & 83.09\% & 61.70\% & 99.73\% \\
    UniFace~\cite{UniFace[20]} & 49.98\% & 50.59\% & 56.34\% & 53.06\% & 94.86\% \\
    E4S~\cite{e4s[54]} & 50.07\% & 49.73\% & 77.61\% & 60.79\% & 93.33\% \\
    DiffSwap~\cite{DiffSwap[55]} & 77.28\% & 52.06\% & 76.95\% & 52.87\% & 93.43\% \\
    FuseAnyPart~\cite{FuseAnyPart[76]} & 84.33\% & 49.99\% & 98.49\% & 52.99\% & 98.53\% \\
    Face-Adapter~\cite{Face-Adapter[77]} & 74.10\% & 50.59\% & 82.54\% & 53.17\% & 87.23\% \\
    REFace~\cite{REFace[78]} & 75.84\% & 70.68\% & 78.33\% & 53.13\% & 94.83\% \\
    \midrule
    Average & 64.04\% & 52.92\% & \underline{75.93\%} & 55.08\% & \textbf{95.24\%} \\
    \bottomrule
  \end{tabular}
  }
  \label{tab:bit_accuracy_celebahq_256}
\end{table}

\subsubsection{Qualitative Evaluation}

We visualized randomly drafted images to validate the visual quality of the watermarked images. As exhibited in Fig.~\ref{fig:common_exhibition}, despite knowing the existence of the watermarks, differences between the raw and watermarked images in the first and second columns, respectively, are visually imperceptible. Starting from column 3, we illustrated the visual effects of the common manipulations on the watermarked images, one for each column. In Fig.~\ref{fig:faceswap_exhibition}, the face swapping effects for the GAN-based\footnote{Diffusion-based face swapping results are omitted since they contain randomness, which naturally leads to slightly different outputs from two identical inferences even without introducing watermarks. } generative models are exhibited in every two columns starting from column 4. The source images that provide the desired synthetic facial identities for each row are placed in the first column. Columns 2 and 3 display the raw and watermarked target images, while every pair of the remaining columns illustrates the face swapping results on the raw and watermarked target images, respectively. It is worth noting that no distortion can be observed in the face swapping results using watermarked target images. In other words, the encoded watermarks do not affect the visual qualities of the synthetic faces, preserving the potential for benign usages of Deepfake face swapping in the industry.


\subsubsection{Quantitative Evaluation} 

In this section, we quantitatively evaluated the visual qualities of watermarked images and the watermark recovery accuracies after common and face swapping manipulations. 

\noindent\textbf{Visual Quality.} We employed the average peak signal-to-noise ratio (PSNR) and structural similarity index measure (SSIM)~\cite{SSIM[27]} between the raw and watermarked images to evaluate the visual quality of watermarked images. As Table~\ref{tab:vis_quality_celeba} reports, while the goal of embedding watermarks into raw images imperceivably stands for all contrastive methods, our proposed approach achieves state-of-the-art visual quality in terms of PSNR and SSIM at both resolution levels, except for ranking a competitive second place for SSIM at the $256 \times 256$ resolution. Meanwhile, for the $128 \times 128$ resolution, RDA~\cite{RootDeepfake[33]} and CIN~\cite{CIN[6]} appear to have reasonable PSNR values above $43$ and SSIM values above $0.95$. Although slightly behind in terms of PSNR, ARWGAN~\cite{ARWGAN[53]} and SepMark~\cite{SepMark[10]} perform well with SSIM values above $0.9$. For the $256 \times 256$ resolution, MBRS~\cite{MBRS[5]} and InvisMark~\cite{InvisMark[75]} are promising for both PSNR and SSIM metrics, where InvisMark exhibits competitive performance compared to our approach. On the other hand, it can be observed that, although pursuing the same objective in visual quality, the proactive distortion models, AntiForgery~\cite{AntiForgery[2]} and CMUA~\cite{CMUA[1]}, have revealed relatively unsatisfactory performance, since noises and perturbations are directly added to raw images, while watermarking approaches execute image reconstructions. 


\noindent\textbf{Watermark Recovery on Common Manipulations.} The watermarks are represented via binary values, and the bit-wise accuracy of the recovered watermark is computed by comparing it to the original watermark following
\begin{equation}
  \textrm{ACC}(m_\textrm{rec}, m) = \frac{l - d_\textrm{ham} (m_\textrm{rec}, m)}{l},
  \label{eq:accuracy}
\end{equation}
where $d_\textrm{ham}$ computes the hamming distance~\cite{Hamming[21]} between the recovered and original watermarks $m_\textrm{rec}$ and $m$, respectively, and $l$ refers to the watermark length. In the experiment, each type of manipulation in $P_\textrm{common}$ and $P_\textrm{swap}$ is performed on the watermarked image $I_\textrm{rec}$ before extracting the watermark via the decoder.  

The watermark recovery performance in the comparative experiment is reported in Table~\ref{tab:bit_accuracy_celebahq_128} and Table~\ref{tab:bit_accuracy_celebahq_256}. Specifically, considering the common manipulations, our method outperforms HiDDeN~\cite{Hidden[4]}, RDA~\cite{RootDeepfake[33]}, CIN~\cite{CIN[6]}, ARWGAN~\cite{ARWGAN[53]}, SepMark~\cite{SepMark[10]}, and WEvade~\cite{jiang2023evading[74]} at the $128 \times 128$ resolution with a better average accuracy, and our performance is competitive against MBRS~\cite{MBRS[5]} with an acceptable average accuracy above $99\%$, achieving the second best statistic. However, it can be observed from Table~\ref{tab:vis_quality_celeba} that MBRS heavily suffers from visual quality issues with unsatisfactory PSNR and SSIM values ($33.01$ and $0.775$) in order to achieve the best bit-wise watermark recovery accuracy in Table~\ref{tab:bit_accuracy_celebahq_128}. As for the $256 \times 256$ resolution, in Table~\ref{tab:bit_accuracy_celebahq_256}, we reached an average accuracy of $99.07\%$ on the common manipulations, outperforming existing state-of-the-art methods. On the contrary, although accomplishes reasonable visual quality as listed in Table~\ref{tab:vis_quality_celeba}, MBRS experiences a large performance damping on the common manipulations in Table~\ref{tab:bit_accuracy_celebahq_256} with respect to the watermark recovery accuracy compared to its $128 \times 128$ resolution version. Meanwhile, although having an average robustness slightly lower than ours, InvisMark~\cite{InvisMark[75]} achieves a promising second-best value of $98.13\%$. On the other hand, IMD~\cite{asnani2022proactive[70]}, MaLP~\cite{asnani2023pro_loc[69]}, and FaceSigns~\cite{FaceSigns[9]} are semi-fragile approaches, where the former two justify watermark recovery performance by the cosine similarity between recovered and original watermarks. Unfortunately, they each fail to stay robust when facing some common manipulations, leading to relatively far gaps from the perfect cosine similarity value ($1.0$) and accuracy ($100.00\%$). More findings are discussed when evaluating them against face swapping manipulations, as they are supposed to be fragile.

\noindent\textbf{Watermark Recovery on Face Swapping.} In this paper, the face swapping manipulation pool contains merely SimSwap~\cite{SimSwap[18]} in the training phase, and we validated the watermark recovery performance for both in- and cross-manipulation settings by introducing GAN-based face swapping algorithms SimSwap~\cite{SimSwap[18]}, InfoSwap~\cite{InfoSwap[19]}, UniFace~\cite{UniFace[20]}, and E4S~\cite{e4s[54]}, and diffusion-based face swapping algorithms DiffSwap~\cite{DiffSwap[55]}, FuseAnyPart~\cite{FuseAnyPart[76]}, Face-Adapter~\cite{Face-Adapter[77]}, and REFace~\cite{REFace[78]} in the testing phase. As shown in Table~\ref{tab:bit_accuracy_celebahq_128} and Table~\ref{tab:bit_accuracy_celebahq_256}, our model consistently recovers the desired watermark messages at average accuracies of $92.50\%$ and $95.24\%$ for $128 \times 128$ and $256 \times 256$ resolutions, respectively, while HiDDeN, MBRS, RDA, CIN, ARWGAN, and InvisMark obtain accuracies around $60\%$ against most synthetic models. WEvade performs better than the former, but still struggles at around $70\%$ on average. Moreover, it can be concluded from the cross-manipulation tests on unseen synthetic models that, by simply tuning with SimSwap, the hidden watermarks are pleasantly maintained by our approach since the goals for different face swapping algorithms are ultimately identical. The semi-fragile watermarking framework, FaceSigns, obtains low watermark recovery accuracies as designed when facing synthetic manipulations that are expected to destroy the embedded watermarks. However, its performance regarding the common manipulations is unexpectedly unsatisfactory besides illustrating low visual quality as reported in Table~\ref{tab:vis_quality_celeba}, leading to an average recovery accuracy of $80.39\%$. Meanwhile, IMD and MaLP exhibit rare robustness when facing one or two particular face swapping algorithms, while having even higher cosine similarities (e.g., $0.7228$ against SimSwap for MaLP) than those when facing common manipulations. Lastly, although preserves outstanding robustness against most common manipulations by obtaining $93.77\%$ and $93.22\%$ average accuracies, SepMark fails to recover the original robust watermarks with only $76.83\%$ and $75.93\%$ average accuracies for the two resolutions, respectively.

\noindent\textbf{Discussion.} In general, Salt and Pepper and Gaussian Noise are the most challenging common manipulations that cause fluctuations in the watermark recovery accuracies for most watermarking algorithms as summarized in Table~\ref{tab:bit_accuracy_celebahq_128} and Table~\ref{tab:bit_accuracy_celebahq_256}. This is consistent with the visualization results in Fig.~\ref{fig:common_exhibition} (col. 6 \& 9) such that these two manipulations add a plethora of perceivable noises to the images and thus heavily jeopardize the hidden watermarks. On the other hand, the contrastive robust watermarking frameworks fail to consistently maintain robustness when facing state-of-the-art face swapping models. As a result, our proposed model is the only one that maintains the bit-wise watermark recovery accuracies above 95\% for all common manipulations and above $80\%$ and $85\%$ at $128 \times 128$ and $256 \times 256$ resolutions, respectively, for all face swapping manipulations, while the contrastive methods are heavily affected by at least one of the manipulations. 

Moreover, although successfully maintains reasonable robustness in general, our method is observed to be affected more when facing the diffusion-based algorithms, with as low as $80.10\%$ regarding robustness against DiffSwap~\cite{DiffSwap[55]}. This performance damping is caused due to the way that diffusion models generate outputs. In particular, DiffSwap edits the target image in a fixed number of iterations, where each iteration derives an intermediate result, leading to possible information loss regarding the embedded watermark. Consequently, the ultimate face swapping results suffer certain levels of damage in the underlying watermarks, which explains the statistical results in Table~\ref{tab:bit_accuracy_celebahq_128} and Table~\ref{tab:bit_accuracy_celebahq_256}. Another observation appears to be that the watermarking algorithms generally perform better at the higher resolution against diffusion-based face swapping models. This is possibly because diffusion-based models are mostly designed to deal with higher resolutions. Meanwhile, some diffusion-based models edit specific facial parts rather than whole face reconstruction, which then leads to easier conditions for watermarks to survive. For instance, FuseAnyPart~\cite{FuseAnyPart[76]} crops and substitutes only the eyes, nose, and mouth areas during the face swapping process, allowing all watermarking algorithms to perform better than when they defend against other diffusion-based models. As a result, despite oscillations in the accuracies against different face swapping models, satisfactory watermark robustness can still be observed for our approach compared to the state-of-the-art ones.

\begin{table}[t!]
  \centering
  \caption{Quantitative visual quality evaluation of the watermarked images on LFW. Information includes model name, image resolution, watermark length, PSNR, and SSIM. The best performance for each resolution is in \textbf{bold} and the second best performance is marked with \underline{underscore}.}
  \begin{tabular}{@{}lcccc@{}}
    \toprule
    Model & Resolution & Length & PSNR$\uparrow$ & SSIM$\uparrow$ \\
    \midrule
    HiDDeN~\cite{Hidden[4]} & $128 \times 128$ & 30 & 33.27 & 0.883 \\
    MBRS~\cite{MBRS[5]} & $128 \times 128$ & 30 & 32.78 & 0.761 \\
    RDA~\cite{RootDeepfake[33]} & $128 \times 128$ & 100 & 42.71 & 0.959 \\
    CIN~\cite{CIN[6]} & $128 \times 128$ & 30 & \underline{42.89} & \underline{0.982} \\
    IMD~\cite{asnani2022proactive[70]} & $128 \times 128$ & -- & 39.31 & 0.918 \\
    MaLP~\cite{asnani2023pro_loc[69]} & $128 \times 128$ & -- & 39.38 & 0.921 \\
    ARWGAN~\cite{ARWGAN[53]} & $128 \times 128$ & 30 & 39.94 & 0.926 \\
    SepMark~\cite{SepMark[10]} & $128 \times 128$ & 30 & 37.03 & 0.947 \\
    WEvade~\cite{jiang2023evading[74]} & $128 \times 128$ & 30 & 32.96 & 0.839  \\
    Ours & $128 \times 128$ & 64 & \textbf{46.00} & \textbf{0.993} \\
    \midrule
    MBRS~\cite{MBRS[5]} & $256 \times 256$ & 256 & \underline{45.37} & 0.969 \\
    FaceSigns~\cite{FaceSigns[9]} & $256 \times 256$ & 128 & 38.23 & 0.910 \\
    SepMark~\cite{SepMark[10]} & $256 \times 256$ & 30 & 38.22 & 0.935 \\
    InvisMark~\cite{InvisMark[75]} & $256 \times 256$ & 100 & 43.60 & \textbf{0.996} \\
    Ours & $256 \times 256$ & 128 & \textbf{45.42} & \underline{0.994} \\
    \bottomrule
  \end{tabular}
  \label{tab:vis_quality_lfw}
\end{table}

\begin{table*}[t!]
  \centering
  \caption{Quantitative comparison on LFW at the $128 \times 128$ resolution regarding the bit-wise recovery accuracy of the watermarks under common manipulations and malicious face swapping manipulations. The best performance for each resolution is in \textbf{bold} and the second best performance is marked with \underline{underscore}.}
  \resizebox{\textwidth}{!}{
  \begin{threeparttable}
  \begin{tabular}{@{}l|cccccccccc@{}}
    \toprule
    Manipulations & HiDDeN~\cite{Hidden[4]} & MBRS~\cite{MBRS[5]} & RDA~\cite{RootDeepfake[33]} & CIN~\cite{CIN[6]} & IMD$^\dagger$~\cite{asnani2022proactive[70]} & MaLP$^\dagger$~\cite{asnani2023pro_loc[69]} & ARWGAN~\cite{ARWGAN[53]} & SepMark~\cite{SepMark[10]} & WEvade~\cite{jiang2023evading[74]} & Ours \\
    \midrule
    Dropout & 82.84\% & 99.99\% & 89.72\% & 99.99\% & 0.5833 & 0.4785 & 97.28\% & 99.98\% & 76.81\% & 99.96\% \\
    Resize & 81.97\% & 99.99\% & 99.89\% & 99.98\% & 0.4039 & 0.4828 & 95.07\% & 99.93\% & 68.59\% & 99.99\% \\
    Jpeg & 56.17\% & 99.42\% & 66.77\% & 97.06\% & 0.5088 & 0.1273 & 70.86\% & 83.42\% & 68.63\% & 99.38\% \\
    GaussianNoise & 51.30\% & 99.80\% & 60.83\% & 87.54\% & 0.4453 & 0.4704 & 53.11\% & 81.90\% & 68.98\% & 99.63\% \\
    SaltPepper & 52.93\% & 99.63\% & 67.31\% & 91.71\% & 0.3789 & 0.2944 & 62.96\% & 99.78\% & 56.93\% & 95.89\% \\
    GaussianBlur & 72.89\% & 99.99\% & 99.88\% & 99.99\% & 0.3404 & 0.4877 & 86.39\% & 91.97\% & 63.04\% & 99.99\% \\
    MedBlur & 82.35\% & 99.99\% & 99.98\% & 99.98\% & 0.4082 & 0.4802 & 96.93\% & 89.72\% & 69.99\% & 99.99\% \\
    \midrule
    Average & 68.63\% & \textbf{99.83\%} & 83.48\% & 96.61\% & 0.4384 & 0.4030 & 80.37\% & 92.39\% & 67.57\% & \underline{99.26\%} \\
    \midrule
    SimSwap~\cite{SimSwap[18]} & 49.96\% & 49.92\% & 50.04\% & 50.25\% & 0.0024 & 0.7341 & 50.00\% & 57.44\% & 59.12\% &99.37\% \\
    InfoSwap~\cite{InfoSwap[19]} & 50.06\% & 50.04\% & 50.10\% & 50.95\% & 0.5258 & 0.1417 & 50.86\% & 73.98\% & 77.93\% & 98.69\% \\
    UniFace~\cite{UniFace[20]} & 53.74\% & 49.78\% & 70.37\% & 49.92\% & 0.4024 & 0.4944 & 59.16\% & 63.10\% & 51.52\% & 98.48\% \\
    E4S~\cite{e4s[54]} & 49.34\% & 50.08\% & 67.78\% & 50.22\% & 0.0064 & 0.6432 & 48.64\% & 88.14\% & 71.03\% & 97.27\% \\
    DiffSwap~\cite{DiffSwap[55]} & 42.36\% & 43.35\% & 61.94\% & 47.15\% & 0.5763 & 0.5616 & 59.25\% & 74.42\% & 75.12\% & 83.29\% \\
    FuseAnyPart~\cite{FuseAnyPart[76]} & 72.26\% & 89.84\% & 28.22\% & 76.67\% & 0.4635 & 0.6390 & 67.16\% & 90.46\% & 86.94\% & 93.15\% \\
    Face-Adapter~\cite{Face-Adapter[77]} & 75.65\% & 85.01\% & 49.67\% & 75.18\% & 0.4542 & 0.6401 & 76.02\% & 77.74\% & 75.58\% & 88.91\% \\
    REFace~\cite{REFace[78]} & 48.85\% & 87.05\% & 49.10\% & 72.94\% & 0.4685 & 0.5668 & 70.60\% & 83.83\% & 64.96\% & 84.96\% \\
    \midrule
    Average & 55.28\% & 63.13\% & 53.40\% & 59.16\% & 0.3624 & 0.5526 & 60.21\% & \underline{76.14\%} & 70.28\% & \textbf{93.02\%} \\
    \bottomrule
  \end{tabular}
  \begin{tablenotes}
    \small {
    \item{ $\dagger$ IMD and MaLP are specially evaluated via cosine similarity according to the original paper. Higher values denote better watermark robustness. }}
  \end{tablenotes}
  \end{threeparttable}
  }
  \label{tab:bit_accuracy_lfw_128}
\end{table*}

\begin{table}[t!]
  \centering
  \caption{Quantitative comparison on LFW at the $256 \times 256$ resolution regarding the bit-wise recovery accuracy of the watermarks under common manipulations and malicious face swapping manipulations. The best performance for each resolution is in \textbf{bold} and the second best performance is marked with \underline{underscore}.}
  \resizebox{\columnwidth}{!}{
  \begin{tabular}{@{}lccccc@{}}
    \toprule
    Manipulations & MBRS~\cite{MBRS[5]} & FaceSigns~\cite{FaceSigns[9]} & SepMark~\cite{SepMark[10]} & InvisMark~\cite{InvisMark[75]} & Ours \\
    \midrule
    Dropout & 99.30\% & 56.45\% & 99.99\% & 99.86\% & 99.97\% \\
    Resize & 99.99\% & 98.91\% & 99.99\% & 97.09\% & 98.16\% \\
    Jpeg & 99.08\% & 75.64\% & 83.41\% & 52.93\% & 99.99\% \\
    GaussianNoise & 55.61\% & 59.90\% & 81.89\% & 66.51\% & 99.92\% \\
    SaltPepper & 61.15\% & 67.99\% & 99.99\% & 56.29\% & 95.99\% \\
    GaussianBlur & 73.93\% & 99.01\% & 91.96\% & 99.99\% & 97.11\% \\
    MedBlur & 99.38\% & 99.91\% & 89.71\% & 99.99\% & 98.13\% \\
    \midrule
    Average & 84.06\% & 79.69\% & \underline{92.42\%} & 81.81\% & \textbf{98.47\%} \\
    \midrule
    SimSwap~\cite{SimSwap[18]} & 50.04\% & 49.77\% & 52.94\% & 52.86\% & 99.96\% \\
    InfoSwap~\cite{InfoSwap[19]} & 50.23\% & 49.91\% & 80.43\% & 62.63\% & 98.67\% \\
    UniFace~\cite{UniFace[20]} & 50.18\% & 52.25\% & 54.10\% & 53.01\% & 97.06\% \\
    E4S~\cite{e4s[54]} & 49.99\% & 50.15\% & 86.58\% & 74.15\% & 97.00\% \\
    DiffSwap~\cite{DiffSwap[55]} & 77.11\% & 53.26\% & 77.38\% & 52.88\% & 98.59\% \\
    FuseAnyPart~\cite{FuseAnyPart[76]} & 83.01\% & 49.13\% & 95.46\% & 53.63\% & 99.20\% \\
    Face-Adapter~\cite{Face-Adapter[77]}  & 83.94\% & 50.63\% & 83.28\% & 53.11\% & 93.81\% \\
    REFace~\cite{REFace[78]} & 69.11\% & 50.17\% & 70.45\% & 52.87\% & 85.15\% \\
    \midrule
    Average & 64.20\% & 50.66\% & \underline{75.08\%} & 56.89\% & \textbf{96.18\%} \\
    \bottomrule
  \end{tabular}
  }
  \label{tab:bit_accuracy_lfw_256}
\end{table}

\subsection{Cross-Dataset Experiment}
\label{sec:cross_dataset}

We further evaluated our watermarking framework on an unseen dataset, Labeled Faces in the Wild (LFW)~\cite{LFW[17]}, regarding the generalization ability. Performance evaluation with respect to the visual quality is listed in Table~\ref{tab:vis_quality_lfw}. Similar results as in Table~\ref{tab:vis_quality_celeba} can be summarized even on a different dataset. In particular, for the $128 \times 128$ resolution, our approach consistently exhibits substantial performance advantages over the comparative ones. As for the $256 \times 256$ resolution, MBRS~\cite{MBRS[5]}, InvisMark~\cite{InvisMark[75]}, and our approach have executed watermarking with promising visual qualities, while FaceSigns~\cite{FaceSigns[9]} and SepMark~\cite{SepMark[10]} fail to do so. 

Meanwhile, as listed in Table~\ref{tab:bit_accuracy_lfw_128} and Table~\ref{tab:bit_accuracy_lfw_256}, the challenging common manipulations concluded in Section~\ref{sec:exp_celeba} consistently lead to unsatisfactory bit-wise watermark recovery accuracies in this section. Subsequently, although slightly affected by Salt and Pepper ($95.89\%$ and $95.99\%$) and Gaussian Blur ($99.99\%$ and $97.11\%$) at $128 \times 128$ and $256 \times 256$ resolutions, respectively, we successfully maintained state-of-the-art watermark recovery accuracies above $95\%$ for all common manipulations. As for watermark robustness against face swapping models, our approach generally performs consistently compared to that on CelebA-HQ, although slight accuracy drops can be observed on average in this cross-dataset scenario. Similarly, a higher average watermark recovery accuracy is achieved at the $256 \times 256$ resolution than at $128 \times 128$, which is mainly reflected in the diffusion-based models.


\subsection{Watermark Collision Check}
\label{sec:collision_check}

As introduced in Section~\ref{sec:id_watermarks}, the binary identity perceptual watermarks are derived from identity embeddings generated via well-designed face recognition tools\footnote{https://github.com/serengil/deepface}. To validate the claim that the identity perceptual watermarks retain the collision-free characteristic of the original identity embeddings, we performed collision checks at different watermark lengths on CelebA-HQ~\cite{CelebAHQ[15]} and LFW~\cite{LFW[17]}. In specific, for the sets of watermarks $M_i$ and $M_j$ of a pair of arbitrary identities $i$ and $j$, the following statement stands,
\begin{equation}
  \forall x \in M_i \forall y \in M_j (x \neq y).
  \label{eq:collision_check}
\end{equation}
In other words, identical watermarks are not allowed to be derived from images that belong to different identities. 

In this paper, images from CelebA-HQ with 6,217 unique identities are adopted to establish $\phi$ and produce watermarks following the pipeline in Section~\ref{sec:id_watermarks}. Thereafter, the watermarks of images in LFW with 5,749 unseen identities are derived following the fixed rule accordingly. As a result, there is no collision in the experiment for CelebA-HQ and LFW at both $128 \times 128$ and $256 \times 256$ resolutions in our collision check. Meanwhile, upon passing the collision check even though not seeing the identities in LFW when defining $\phi$, it can be concluded that the number of unique identities in CelebA-HQ is sufficient to avoid the potential bias, and the cross-dataset performance is therefore reliably preserved.

\subsection{Deepfake Detection}
\label{sec:deepfake_detection}

In this section, the Deepfake detection task is proactively achieved by evaluating the recovered watermark $m_\textrm{rec}$. Specifically, for a target image $I$ with identity $t$ to be proactively protected with an identity perceptual watermark $m$, the watermarked image $I_\textrm{rec}$ is firstly derived via the encoder of our framework. Then, when a face swapping algorithm is applied on $I_\textrm{rec}$ to generate $I_\textrm{fake}$ with a new identity $s$, an identity perceptual watermark $m_s$ is then derived based on the image content of $I_\textrm{fake}$. Meanwhile, the robust watermark is also recovered, denoted as $m_\textrm{rec}$, via the decoder of our framework. In the end, the two watermarks, $m_s$ and $m_\textrm{rec}$, are compared bit-wisely and the falsification can be addressed based on their matching rate. In other words, since $m_\textrm{rec}$ represents a perfect recovery of the original embedded watermark that corresponds to the original facial identity and $m_s$ represents the watermark that corresponds to the identity in the current image content, a low similarity between the two implies distinct facial identities for images $I_\textrm{rec}$ and $I_\textrm{fake}$, indicating that Deepfake face swapping happens from $I_\textrm{rec}$ to $I_\textrm{fake}$. Contrarily, a promisingly high similarity between $m_s$ and $m_\textrm{rec}$ implies that the facial identity remains unchanged. It is worth noting that the entire proactive Deepfake detection via our proposed identity perceptual watermarks does not require any ground-truth watermarks. 

\begin{table*}[t!]
    \centering
    \caption{Deepfake detection performance in AUC scores against different face swapping algorithms on CelebA-HQ at $128 \times 128$ and $256 \times 256$ resolutions. The last row evaluates the detection performance on a mixture of all face swapping algorithms. The best performance at each resolution is in \textbf{bold}.}
    \begin{tabular}{@{}lcccccccccc@{}}
    \toprule
         & \multicolumn{2}{c}{Xception~\cite{DeepfakeBenchmark[56]}} & \multicolumn{2}{c}{SBIs~\cite{SBIs[57]}} & \multicolumn{2}{c}{RECCE~\cite{RECCE[58]}} & \multicolumn{2}{c}{CADDM~\cite{CADDM[59]}} & \multicolumn{2}{c}{Ours} \\
    \midrule
         Resolution & 128 & 256 & 128 & 256 & 128 & 256 & 128 & 256 & 128 & 256 \\
    \midrule
        SimSwap~\cite{SimSwap[18]} & 98.78\% & \textbf{99.99\%} & 85.38\% & 98.63\% & 99.55\% & 99.98\% & \textbf{99.98\%} & 99.98\% & 98.49\% & 98.19\% \\
        InfoSwap~\cite{InfoSwap[19]} & 33.93\% & 48.98\% & 61.59\% & 81.55\% & 43.06\% & 56.32\% & 45.92\% & 56.56\% & \textbf{99.02\%} & \textbf{99.52\%} \\
        UniFace~\cite{UniFace[20]} & 55.29\% & 46.03\% & 84.71\% & 81.64\% & 78.57\% & 53.79\% & 63.55\% & 64.61\% & \textbf{99.99\%} & \textbf{98.99\%} \\
        E4S~\cite{e4s[54]} & 30.13\% & 39.84\% & 55.64\% & 67.06\% & 26.83\% & 40.81\% & 28.63\% & 36.85\% & \textbf{97.71\%} & \textbf{98.70\%} \\
        DiffSwap~\cite{DiffSwap[55]} & 65.40\% & 53.62\% & 56.71\% & 68.65\% & 86.18\% & 63.60\% & 81.93\% & 73.33\% & \textbf{92.73\%} & \textbf{98.09\%} \\
        FuseAnyPart~\cite{FuseAnyPart[76]} & 31.15\% & 62.50\% & 77.42\% & 79.58\% & 74.71\% & 63.80\% & 74.23\% & 81.34\% & \textbf{93.96\%} & \textbf{98.26\%} \\
        Face-Adapter~\cite{Face-Adapter[77]} & 30.57\% & 62.52\% & 75.72\% & 66.64\% & 73.61\% & 75.51\% & 72.86\% & 67.83\% & \textbf{95.80\%} & \textbf{99.15\%} \\
        REFace~\cite{REFace[78]} & 29.49\% & 31.25\% & 66.51\% & 76.88\% & 65.32\% & 64.12\% & 65.77\% & 65.21\% & \textbf{82.49\%} & \textbf{85.93\%} \\
    \midrule
        Mixed & 33.01\% & 66.89\% & 67.85\% & 75.19\% & 66.61\% & 69.94\% & 75.90\% & 69.25\% & \textbf{95.39\%} & \textbf{97.57\%} \\
    \bottomrule
    \end{tabular}
    \label{tab:deepfake_detection}
\end{table*}

Since existing comparative proactive watermarking approaches do not assign semantics to their watermarks, Deepfake detection cannot be accomplished without knowing the matched watermark for each image in advance, and this is usually unavailable in real-life scenarios. Meanwhile, the unsatisfactory watermark recovery accuracies in Section~\ref{sec:exp_celeba} suggest that evaluating detection performance can be trivial for them. Therefore, in this experiment, we adopted several popular and state-of-the-art passive Deepfake detectors in recent years for comparison, namely, Xception~\cite{FFPP[11]}, SBIs~\cite{SBIs[57]}, RECCE~\cite{RECCE[58]}, and CADDM~\cite{CADDM[59]}. Since our approach is trained on CelebA-HQ with SimSwap accessible, we trained the detectors on CelebA-HQ against SimSwap, rather than directly adopting the model weights trained on other datasets, to guarantee fairness in the experiment. In particular, images in CelebA-HQ are recognized as real samples, and the fake samples are derived by executing face swapping on CelebA-HQ via SimSwap.  

Since the accuracy metric requires a fixed threshold value, which can lead to biases due to data distributions, we employed the area under the receiver operating characteristic (ROC) curve (AUC) score to evaluate the detection performance based on the matching rate. In a nutshell, a high AUC score demonstrates the probability that a random positive sample scores higher than a random negative sample from the testing set, in other words, the ability of the classifier to distinguish between real and fake samples. The experiment is conducted on CelebA-HQ at $128 \times 128$ and $256 \times 256$ resolutions by testing against each face swapping model in $P_\textrm{swap}$ and a mixture\footnote{When conducting this experiment, samples are equally and randomly drawn for each synthetic algorithm to match the same total quantity as the real ones. } of them. To establish a balanced ratio for real and fake samples, the real images in CelebA-HQ are leveraged after randomly applying common manipulations from $P_\textrm{common}$.

The detection performance in AUC scores is demonstrated in Table~\ref{tab:deepfake_detection}. It can be observed that, except SBIs that reconstructs its own fake training samples following the designed pipeline, all models have achieved outstanding AUC scores on detecting SimSwap, where CADDM reaches 99.98\% at the $128 \times 128$ resolution and Xception achieves 99.99\% at the $256 \times 256$ resolution, slightly outperforming our promising performance of 98.49\% and 98.19\% at the two resolutions, respectively. This demonstrates the reliable detection ability of the passive detectors on seen manipulations. However, when facing unseen face swapping manipulations, the passive detectors generally come up with poor statistics of at most 86.18\%. At the same time, many statistics are observed below 50\% regarding the passive detectors, demonstrating substantial fluctuations in the detection performance against different synthetic models. For instance, RECCE achieves an 86.18\% AUC score against DiffSwap at the $128 \times 128$ resolution, but a poor 26.83\% statistic is derived against E4S. As a result, huge performance damping can be concluded for the passive detectors since the face swapping models are unseen and can generate faces with imperceptible detectable artifacts. 

Contrarily, our proposed approach successfully retains promising AUC scores above $90\%$ at both resolutions for both seen and unseen face swapping manipulations except for REFace. This implies that the matching rates between $m_s$ and $m_\textrm{rec}$ for images after common manipulations (real) are consistently higher than those for images after face swapping manipulations (fake). In other words, our identity perceptual watermarking framework not only robustly survives image manipulations, but also favorably distinguishes real and fake samples with respect to the changes in facial identities. Meanwhile, REFace has brought the most challenges to our watermarks, although the watermark robustness is proved to be good in Table~II and Table~III ($94.61\%$ and $94.83\%$). This reflects the potential doubt that REFace generates faces that still contain similar identities as the original ones, making higher similarity between $m_s$ and $m_\textrm{rec}$\footnote{Since this refers to unsatisfactory face swapping performance, there is no route to improve the proposed watermarks accordingly.}. As a result, for a mixture of all eight Deepfake models, including seven unseen ones, our approach retains satisfactory detection performance at $95.39\%$ and $97.57\%$ at $128 \times 128$ and $256 \times 256$ resolutions, respectively, against both GAN-based and diffusion-based Deepfake face swapping models, demonstrating superior generalization ability.

\begin{table}[t!]
\centering
\caption{Top-N source tracing accuracy of watermarks on CelebA-HQ. }
\resizebox{\columnwidth}{!}{
\begin{tabular}{lcccccc}
\toprule
\multirow{2}{*}{Model} & \multicolumn{2}{c}{Acc@5} & \multicolumn{2}{c}{Acc@3} & \multicolumn{2}{c}{Acc@1} \\
\cmidrule{2-7}
 & SepMark~\cite{SepMark[10]} & Ours & SepMark~\cite{SepMark[10]} & Ours & SepMark~\cite{SepMark[10]} & Ours \\
\midrule
SimSwap~\cite{SimSwap[18]} & 0.25\% & 99.96\% & 0.10\% & 99.96\% & 0.03\% & 99.89\% \\
InfoSwap~\cite{InfoSwap[19]} & 5.68\% & 99.99\% & 4.08\% & 99.99\% & 2.27\% & 99.92\% \\
UniFace~\cite{UniFace[20]} & 16.02\% & 99.43\% & 13.71\% & 99.36\% & 9.38\% & 98.72\% \\
E4S~\cite{e4s[54]} & 69.63\% & 99.43\% & 68.57\% & 99.04\% & 66.12\% & 98.08\% \\
DiffSwap~\cite{DiffSwap[55]} & 1.33\% & 99.75\% & 0.97\% & 99.67\% & 0.37\% & 99.31\% \\
FuseAnyPart~\cite{FuseAnyPart[76]} & 1.28\% & 99.99\% & 1.08\% & 99.96\% & 0.30\% & 99.87\% \\
Face-Adapter~\cite{Face-Adapter[77]} & 1.85\% & 86.47\% & 1.48\% & 84.03\% & 0.37\% & 77.36\% \\
REFace~\cite{REFace[78]} & 4.69\% & 99.79\% & 2.34\% & 99.72\% & 0.80\% & 99.33\% \\
\midrule
Average & 12.59\% & 98.10\% & 11.54\% & 97.72\% & 9.96\% & 96.56\% \\
\bottomrule
\end{tabular}
}
\label{tab:source_tracing}
\end{table}

\begin{table*}[t!]
  \centering
  \caption{Robustness ablation test on common manipulations for easier and harder parameter values that are unseen during training. The bit-wise watermark recovery accuracies are reported. Long common manipulation names are abbreviated.}
  \begin{tabular}{@{}lcccccccccccccc@{}}
    \toprule
    \multirow{2}{*}{Difficulty} & \multicolumn{2}{c}{Dropout ($r$)} & \multicolumn{2}{c}{Resize ($r$)} & \multicolumn{2}{c}{Jpeg ($Q$)} & \multicolumn{2}{c}{GNoise ($\mu$, $\sigma$)} & \multicolumn{2}{c}{S\&P ($r$)} & \multicolumn{2}{c}{GBlur ($\sigma$, $k$)} & \multicolumn{2}{c}{MBlur ($k$)} \\
    \cmidrule{2-15}
    & $r$ & Acc & $r$ & Acc & $Q$ & Acc & $\mu$, $\sigma$ & Acc & $r$ & Acc & $\sigma$, $k$ & Acc & $k$ & Acc \vspace*{-0.05cm}\\
    \midrule
    Easy & 0.3 & 99.99\% & 0.3 & 99.99\% & 70 & 99.99\% & 0, 0.05 & 99.99\% & 0.03 & 99.76\% & 1, 1 & 99.99\% & 1 & 99.99\% \\
    Regular & 0.5 & 99.99\% & 0.5 & 99.99\% & 50 & 99.48\% & 0, 0.1 & 98.63\% & 0.05 & 95.54\% & 2, 3 & 99.99\% & 3 & 99.99\% \\
    Hard & 0.7 & 99.46\% & 0.7 & 99.82\% & 30 & 99.38\% & 0, 0.2 & 93.24\% & 0.07 & 94.19\% & 3, 5 & 98.08\% & 5 & 99.50\% \\
    \bottomrule
  \end{tabular}
  \label{tab:ablation}
\end{table*}

\subsection{Source Tracing}
\label{sec:source_tracing}

In this paper, the proposed identity perceptual watermark aims to perform Deepfake detection and source tracing concurrently based on its robustness. Therefore, besides demonstrating promising watermark recovery accuracies and Deepfake detection AUC scores in early sections, in this section, we evaluated the watermark ability on source tracing. In particular, we treated each identity as a unique class and evaluated the top-5 (ACC@5), top-3 (ACC@3), and top-1 (ACC@1) accuracies on the recovered watermarks with respect to the images after face swapping. In other words, for each recovered identity perceptual watermark, we matched it back to the original facial identities and recorded the percentage of cases such that the correct identity lies in the top-5, top-3, and top-1 watermark similarities. For a fair comparison, we selected the state-of-the-art watermarking framework with the most competitive robustness in terms of watermark recovery accuracy, SepMark, as the contrastive method in the source tracing experiment. 

The source tracing performance of the watermarks is reported in Table~\ref{tab:source_tracing}. It can be observed that our approach promisingly preserves the top-5 accuracies above $99\%$ against all face swapping algorithms except for Face-Adapter. Meanwhile, even as the rule becomes more strict for top-3 and top-1 accuracies, the statistics are still maintained above $98\%$. As for the challenging Face-Adapter, the results are aligned with Table~III where it causes a relatively lower watermark recovery accuracy of $87.23\%$. On the other hand, the promising source tracing performance against REFace further proves the fact that the Deepfake detection performance in Table~VII is caused by the unsatisfactory generative performance of REFace. Contrarily, except performing $69.93\%$, $68.57\%$, and $66.12\%$ accuracies for top-5, top-3, and top-1 against E4S, the opponent that SepMark gains the best watermark robustness, SepMark derives generally poor accuracies below $20\%$ for all other face swapping models. As a result, our approach has been proved to have outstanding source tracing ability to find the original victim facial identities by providing reliable top-5, top-3, and top-1 accuracies regarding the face swapping manipulations.


\begin{table}[t!]
\centering
\caption{Demonstration of the cyclic oscillations of $x_i$ with $x_0=0.1$ for different $r$ values. }
\begin{tabular}{lcccc}
\toprule
$x_0=0.1$ & $r = 2$ & $r = 2.5$ & $r = 3.5$ & $r = 3.93$ \\
\midrule
$x_1$ & 0.18 & 0.225 & 0.315 & 0.3537 \\
$x_2$ & 0.2952 & 0.435938 & 0.755213 & 0.898383 \\
$x_3$ & 0.416114 & 0.614740 & 0.647033 & 0.358772 \\
$x_4$ & 0.485926 & 0.592086 & 0.799334 & 0.904114 \\
$x_5$ & 0.499603 & 0.603800 & 0.561395 & 0.340696 \\
\ldots & \ldots & \ldots & \ldots & \ldots \\
$x_{10}$ & \textcolor{red}{0.5} & \textcolor{red}{0.599880} & \textcolor{red}{0.864143} & 0.611306 \\
$x_{11}$ & \textcolor{red}{0.5} & \textcolor{blue}{0.600060} & \textcolor{blue}{0.410898} & 0.933811 \\
$x_{12}$ & 0.5 & \textcolor{red}{0.599970} & \textcolor{green}{0.847213} & 0.242905 \\
$x_{13}$ & 0.5 & \textcolor{blue}{0.600015} & \textcolor{brown}{0.453051} & 0.722736 \\
$x_{14}$ & 0.5 & 0.599992 & \textcolor{red}{0.867285} & 0.787528 \\
$x_{15}$ & 0.5 & 0.600003 & \textcolor{blue}{0.402856} & 0.657598 \\
$x_{16}$ & 0.5 & 0.599998 & \textcolor{green}{0.841970} & 0.884890 \\
$x_{17}$ & 0.5 & 0.600001 & \textcolor{brown}{0.465697} & 0.400310 \\
$x_{18}$ & 0.5 & 0.599999 & 0.870882 & 0.943443 \\
$x_{19}$ & 0.5 & 0.600000 & 0.393564 & 0.209698 \\
$x_{20}$ & 0.5 & 0.599999 & 0.835350 & 0.651298 \\
\bottomrule
\end{tabular}
\label{tab:chaotic_table}
\end{table}

\subsection{Ablation Studies}
\label{sec:abalation_studies}

\noindent \textbf{Common Manipulations at Various Difficulties.} In this paper, following the convention of steganography, we picked the common manipulations that best imitate the unavoidable image quality degradation of posting and spreading in real-life scenarios and set the parameter values with considerable difficulties. Particularly, the parameters are set equal to or harder than the early work~\cite{Hidden[4],MBRS[5],CIN[6]} in our experiments. In this section, to further validate the robustness of the proposed watermarking framework, we conducted ablation testing sessions on CelebA-HQ at the $256 \times 256$ resolution with easier and harder parameters that are unseen during training for the common manipulations, respectively. As a result, in addition to the outstanding performance reported in Sections~\ref{sec:exp_celeba} and~\ref{sec:cross_dataset}, Table~\ref{tab:ablation} implies that the robustness is completely guaranteed for common manipulations with easier parameter values, and it is also only slightly affected in harder conditions, with all accuracies still above $90\%$. In conclusion, the potential effects of common manipulations at different difficulty levels are trivial with respect to the robustness of our proposed approach. 

\noindent \textbf{Watermark Confidentiality Analyses.} A chaotic encryption system is devised to securely protect the watermarks in Section~\ref{sec:id_watermarks} to avoid potential adaptive attacks. In particular, an attacker may attempt to reversely deduce the original watermark based on an encrypted one. In our approach, following the parameter settings in Section~\ref{sec:impl_detail}, the encryption pipeline becomes unpredictable and nonreversible relying on Eqn.($\ref{eq:logistic_map}$). In Table~\ref{tab:chaotic_table}, besides fixing $x_0$ at the chosen value $0.1$ in this study, we listed the sequence of $x_i$ for $1 <= i <= 30$ by attempting different $r$ values. It can be observed that, after a few iterations, $x_i$ starts to oscillate between cyclic values for $r = 2$, $r = 2.5$, and $r = 3.5$. Specifically, $x_i$ stables at $0.5$ (red) for $r = 2$ and varies around $0.59$ (red) and $0.60$ (blue) for $r = 2.5$. As for $r = 3.5$, four oscillation patterns can be concluded between approximately $0.86$ (red), $0.40$ (blue), $0.84$ (green), and $0.46$ (brown). On the contrary, the evolution of $x_i$ is completely unpredictable with no patterns observed for the chosen $r = 3.93$ in this study. In other words, with no evolution pattern in the chaotic sequence, the adaptive attacks are prevented. 

\begin{figure}[t!]
  \centering
   \includegraphics[width=\columnwidth]{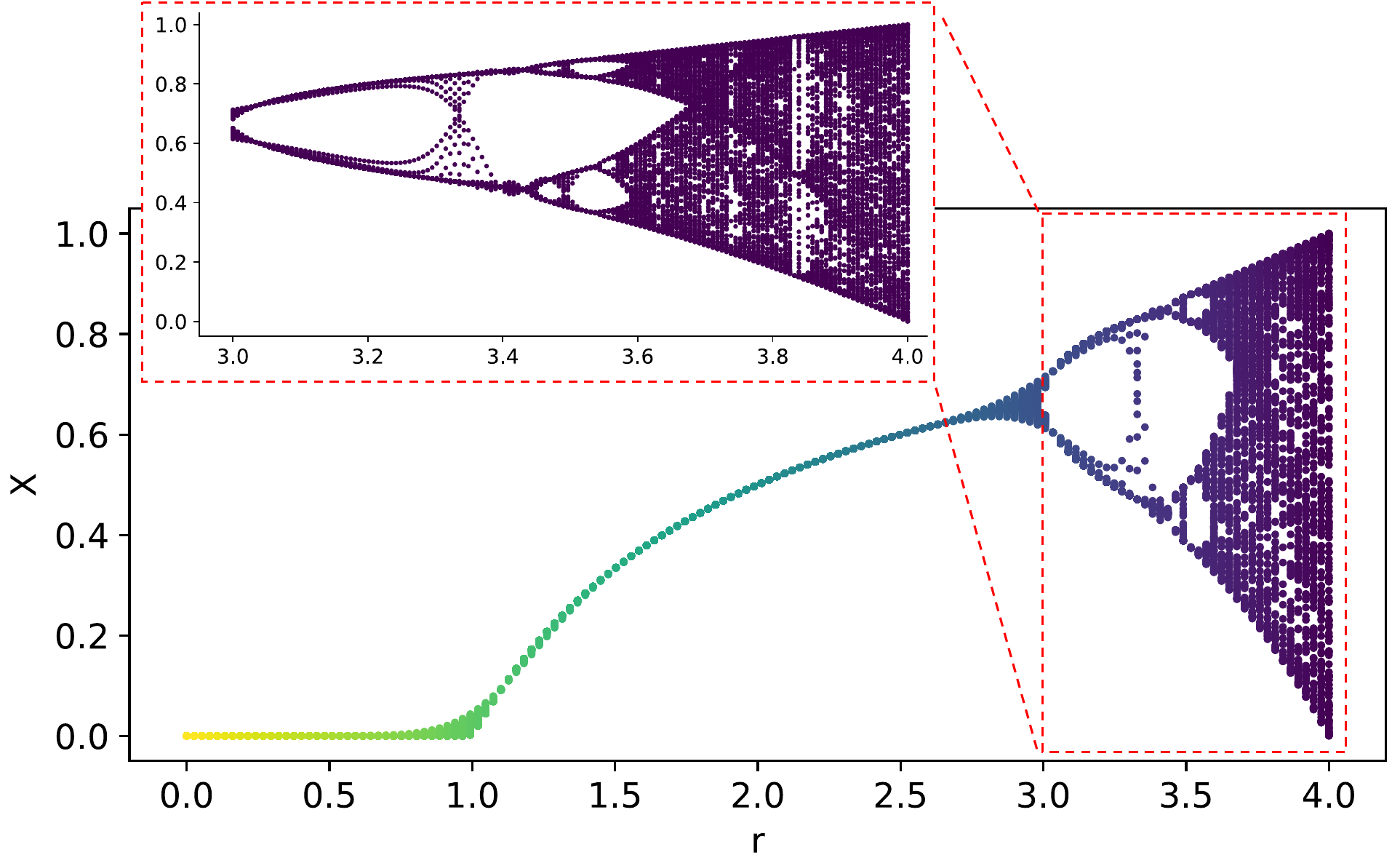}
   \caption{Bifurcation diagram of the logistic map Eqn.($\ref{eq:logistic_map}$) for continuous $r$ values with $x_0 = 0.1$. }
   \label{fig:chaotic_map}
\end{figure}

\begin{figure}[t!]
  \centering
   \includegraphics[width=\columnwidth]{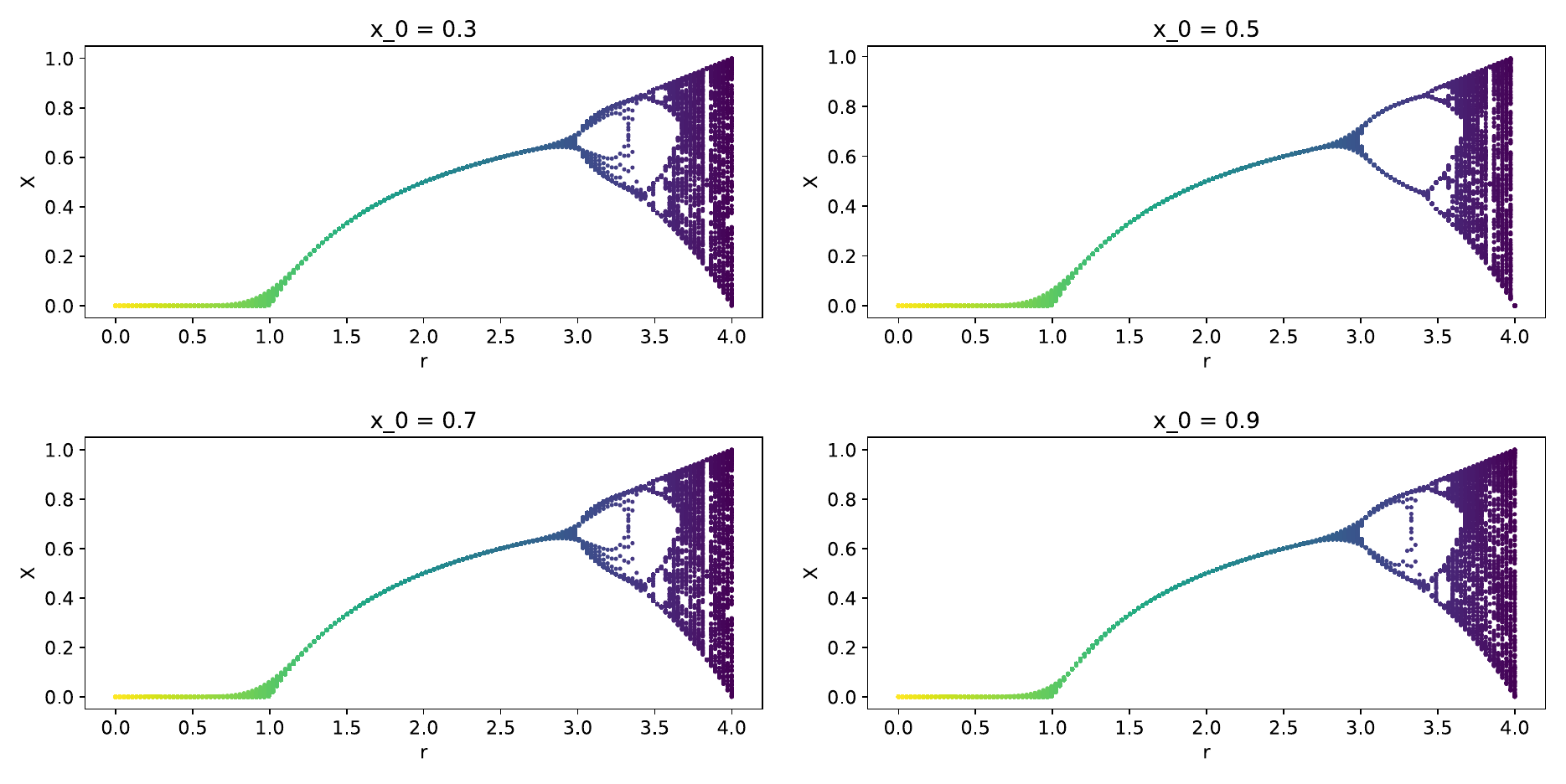}
   \caption{Bifurcation diagrams with different $x_0$ values. }
   \label{fig:chaotic_ablation}
\end{figure}

In Fig.~\ref{fig:chaotic_map}, we visualized the bifurcation diagram of the logistic map of Eqn.($\ref{eq:logistic_map}$) with $x_0 = 0.1$ for a continuous range of $0 <= r <= 4$. In particular, by omitting $x_i$ for $0 <= i < 10$, the $200$ consecutive $x_i$ values are plotted as points for each $r$ value after deriving $150$ linear spaces within the range. Moreover, we zoomed in on the view to $3 <= r <= 4$ for a closer observation of the oscillation and unpredictable behaviors of $x_i$ at different $r$ values. It can be concluded that the chaotic characteristic in $X$ occurs for approximately $r > 3.6$ after periodic sequences for lower $r$ values. 

Besides, we presented the visualizations with different $x_0$ values in Fig.~\ref{fig:chaotic_ablation}, which proves that the choice of $x_0$ does not affect the chaotic behavior. Therefore, selecting a proper $r$ value with a chaotic sequence $X$ for the generation of the encryption key following Eqn.($\ref{eq:key_encrypt}$) makes the watermark encryption pipeline unpredictable and nonreversible even though the adversarial attackers may have access to the encrypted watermarks. Meanwhile, it is not necessary to adopt the first $l$ values in $X$ for watermark encryption, because randomly selecting an $x_i$ for $i \neq 0$ as the starting point to derive the encryption key $K$ can further ensure watermark safety.

Furthermore, we provided a demonstration of encrypting a sample binary watermark $m_c$ of length 20 using the chaotic encryption system following Eqn.($\ref{eq:logistic_map}$) and Eqn.($\ref{eq:key_encrypt}$). To begin with, we defined the sample watermark as follows.
\begin{lstlisting}
$m_c$: 0 1 0 1 0 1 0 0 0 1 1 1 0 0 0 1 0 1 0 0
\end{lstlisting}
To encrypt $m_c$, we adopted $x_{10}$ through $x_{19}$ with $r= 3.93$ from Table~\ref{tab:chaotic_table} to generate the encryption key sequence $K$. Following Eqn.($\ref{eq:key_encrypt}$) with constant values $p=5$ and $q=11$, the $x_i$ values are transformed to $K$ as follows.
\begin{lstlisting}
$K$: 0 1 0 0 1 1 1 0 0 0 1 1 1 1 1 1 0 1 0 0
\end{lstlisting}
An XOR operation between $m_c$ and $K$ derives $m_e$ as follows.
\begin{lstlisting}
$m_e$: 0 0 0 1 1 0 1 0 0 1 0 0 1 1 1 0 0 0 0 0
\end{lstlisting}
This encrypted watermark $m_e$ is confidentially protected against adversarial attackers and can be used for watermark embedding. Contrarily, an authorized individual can easily decrypt $m_e$ to derive $m_d$ by an extra XOR operation with the encryption key $K$.
\begin{lstlisting}
$m_d$: 0 1 0 1 0 1 0 0 0 1 1 1 0 0 0 1 0 1 0 0
\end{lstlisting}
As a result, one can find it faithfully identical to the original watermark $m_c$. 

\section{Conclusion}

In this paper, we propose a robust identity perceptual watermarking framework that proactively defends against malicious Deepfake face swapping. By innovatively and securely assigning identity semantics to the watermarks, our approach concurrently accomplishes Deepfake detection and source tracing with a single robust watermark for the first time. Meanwhile, confidentiality is guaranteed by devising a chaotic encryption system that brings unpredictability to identity perceptual watermarks. Extensive experimental results have proved that our method is robust when faced against both common and malicious manipulations, even under cross-dataset settings. Moreover, by leveraging a single face swapping algorithm, SimSwap, for fine-tuning, we are able to preserve promising cross-manipulation performance against unseen face swapping algorithms. The AUC scores of $95.39\%$ and $97.57\%$ at $128 \times 128$ and $256 \times 256$ resolutions against mixed manipulations have further proved the promising detection ability of our approach against Deepfake face swapping, significantly outperforming the state-of-the-art passive detectors. Lastly, the reliable top-5, top-3, and top-1 source tracing accuracies of 98.10\%, 97.72\%, and 96.56\% favorably demonstrate the ability of our method to trace back to the original victim identity along with proactive Deepfake detection. 

Admittedly, since the watermarks are identity perceptual, the proposed framework is unable to detect face re-enactment manipulations that modify facial expressions and poses but retain facial identities. Our future work will focus on designing a unified framework to proactively defend against both face swapping and face re-enactment manipulations by analyzing and introducing the corresponding facial attribute features. 


 
\bibliography{IEEEabrv,my_bib}
\bibliographystyle{IEEEtran}

\newpage

\begin{IEEEbiography}[{\includegraphics[width=1in,height=1.25in,clip,keepaspectratio]{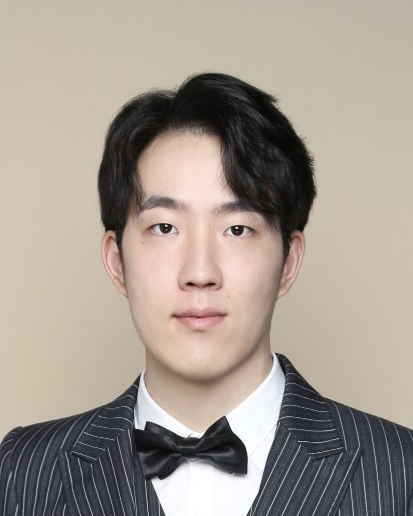}}]{Tianyi Wang}
(Member, IEEE) received the double major B.S. degrees in Computer Science and Applied and Computational Mathematical Sciences from the University of Washington, Seattle, USA, in 2018. After that, he received the Ph.D. degree in Computer Science, under the supervision of Dr. Kam Pui Chow, from The University of Hong Kong, Hong Kong, in 2023. He is currently a Postdoctoral Research Fellow at the School of Computing, National University of Singapore, Singapore. His major research interests include multimedia forensics, misinformation analysis, face forgery detection, generative artificial intelligence, and computer vision. 
\end{IEEEbiography}

\begin{IEEEbiography}
[{\includegraphics[width=1in,height=1.25in,clip,keepaspectratio]{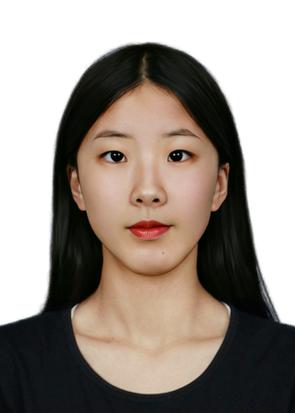}}] {Mengxiao Huang}
graduated from Zaozhuang University in 2018 with a bachelor's degree and is currently studying for a master's degree in mathematics and artificial intelligence at Qilu University of Technology (Shandong Academy of Sciences). Her research interests include Deepfake detection, active defense, and computer vision.
\end{IEEEbiography}

\begin{IEEEbiography}
[{\includegraphics[width=1in,height=1.25in,clip,keepaspectratio]{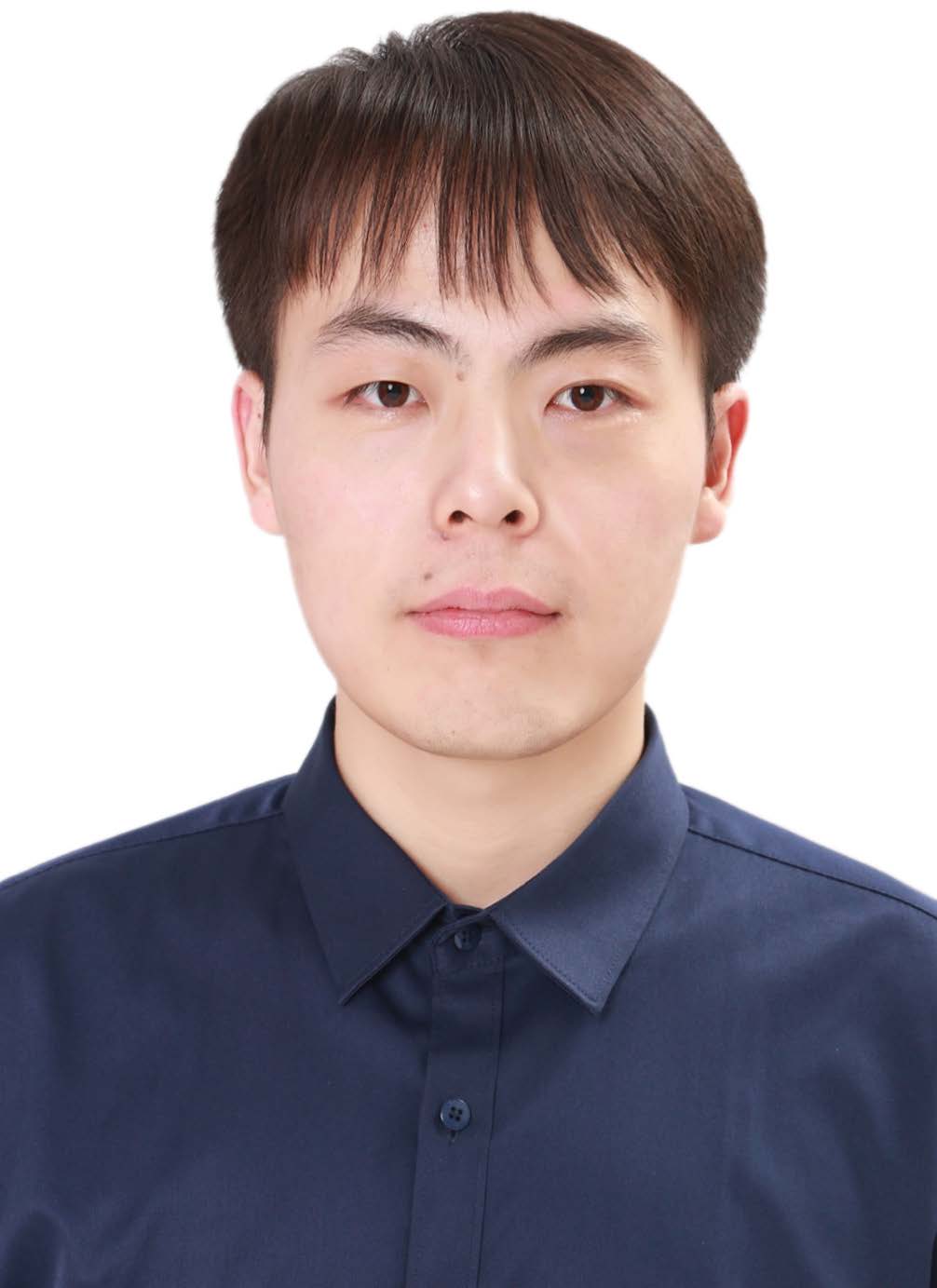}}] {Harry Cheng}
(Member, IEEE) is a Research Fellow at the National University of Singapore, Singapore. He received his Ph.D. degree from Shandong University, China. He has authored or coauthored several papers in top conferences and journals, including NeurIPS, ICCV, ACM MM, and IEEE TMM, and he serves as a regular reviewer for several journals such as IEEE TIP, IEEE TIFS, IEEE TKDE, IEEE TMM, and ACM ToMM.
\end{IEEEbiography}

\begin{IEEEbiography}
[{\includegraphics[width=1in,height=1.25in,clip,keepaspectratio]{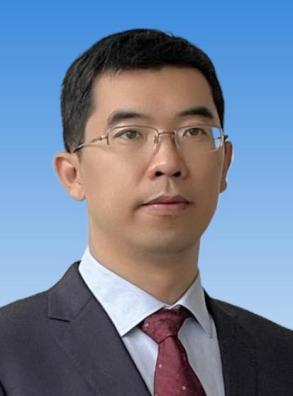}}] {Bin Ma}
(Member, IEEE) received the M.S. and Ph.D. degrees from Shandong University, Jinan, China, in 2005 and 2008, respectively. From 2008 to 2013, he was an Associate Professor with the School of Information Science, Shandong University of Political Science and Law, Jinan. He visited the New Jersey Institute of Technology at Newark, NJ, USA, as a Visiting Scholar, from 2013 to 2015. He is currently a Professor with the School of Cyber Security, Qilu University of Technology (Shandong Academy of Sciences), Shandong, China. He is also with the Shandong Provincial Key Laboratory of Computer Networks, Jinan. His research interests include reversible data hiding, multimedia security, and image processing. He serves as an Editorial Board Member for a few journals, such as the IEEE Transactions on Information Forensics and Security, the Journal of Visual Communication and Image Representation, and the IEEE Signal Processing. He is a member of ACM.
\end{IEEEbiography}

\begin{IEEEbiography}[{\includegraphics[width=1in,height=1.25in,clip,keepaspectratio]{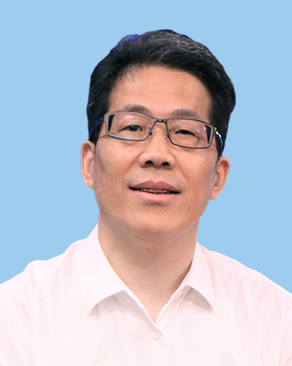}}] {Yinglong Wang}
(Senior Member, IEEE) is currently a Researcher, Ph.D. Supervisor. He is the Director of the Key Laboratory of Computing Power Network and Information Security, Ministry of Education. In recent years, he has taken charge of more than 20 national, provincial, and ministerial projects. He has authored or coauthored more than 100 top academic articles and owns more than 120 authorized invention patents. His main research interests include high-performance computing, information security, and medical artificial intelligence. He is granted as a Young and Middle-aged Expert with outstanding contributions to Shandong Province, Shandong province Taishan scholar climbing plan expert, High-End Think Tank Expert of Shandong Province, and enjoys special government allowances from the State Council. He is the President of the Shandong Artificial Intelligence Association. The scientific research projects led by him won three First Prizes of the Shandong Science and Technology Progress Award.
\end{IEEEbiography}

\vspace{11pt}

\vfill

\end{document}